\documentclass[10pt,twocolumn,letterpaper]{article}
\usepackage[pagenumbers]{cvpr}

\usepackage{cuted}      
\usepackage{currfile}
\usepackage{caption}
\usepackage{multirow}
\usepackage{threeparttable}
\usepackage{xspace}
\usepackage{microtype}
\usepackage{enumitem}
\usepackage{array}
\newcolumntype{P}[1]{>{\raggedright\arraybackslash}p{#1}}   

\DeclareMathOperator*{\argmin}{arg\,min}
\newtheorem{proposition}{Proposition}

\newcommand{\dens}{b}                       
\newcommand{\lay}{\ell}                     
\newcommand{\eprobe}{\hat e}                
\newcommand{\lut}{r}                        

\newcommand{\ttiny}{\scriptsize}
\AtBeginDocument{\crefname{figure}{Figure}{Figures}}

\AtBeginDocument{\raggedbottom}

\definecolor{cvprblue}{rgb}{0.21,0.49,0.74}
\usepackage[breaklinks,colorlinks,allcolors=cvprblue]{hyperref}

\def\confName{CVPR}
\def\confYear{2027}

\title{Sparsity Should Be ALLOTted by Output: Training-Free Allocation of Block-Sparse Attention for Video Diffusion Transformers}

\author{\textbf{Tianze Xia}$^{1,2,4}$\thanks{Work done during an internship at KlingAI.\quad $^\dagger$Corresponding author.}\quad
\textbf{Xiaoling Zhou}$^{3,4}$\quad
\textbf{Ruya Lan}$^{1}$\quad
\textbf{Yiqiao Liao}$^{4}$\quad
\textbf{Chengru Song}$^{4}$\quad
\textbf{Shiming Xiang}$^{1,2\dagger}$\\
$^1$UCAS\quad
$^2$CASIA\quad
$^3$PKU\quad
$^4$KlingAI Research\\
}

\begin{document}
\maketitle
\begin{figure*}[t]
\centering
\includegraphics[width=\textwidth]{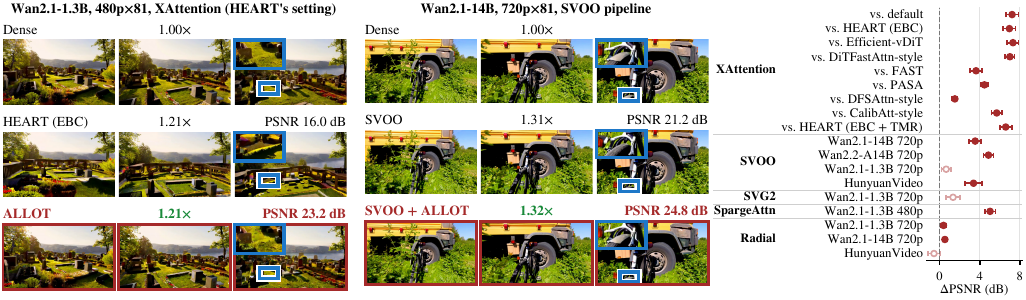}
\caption{\textbf{Changing only the allocation at equal wall-clock time.}
Left: dense, baseline and +\,ALLOT (top to bottom), with blue-box insets; prompts are near the median gain.
Right: paired PSNR gains (ALLOT $-$ baseline) with 95\% CIs for \cref{tab:compare,tab:plugin}; open markers denote ties.}
\label{fig:teaser}
\end{figure*}

\begin{abstract}
Full 3D attention is costly in video Diffusion Transformers.
Block-sparse attention reduces this cost by computing only selected attention blocks.
{However, existing methods either overlook the potential to optimize sparsity allocation or rely on heuristic rules or local metrics to guide it.
In this work, we observe that sparsifying attention in different layers and at different denoising steps affects final-video fidelity to substantially different degrees. Yet the identities of the most sensitive layer–step pairs remain nearly unchanged across inputs.
Building on these findings, we propose \textbf{A}llocating \textbf{L}ayer--step \textbf{L}atency by \textbf{O}utput \textbf{T}ests (\textbf{ALLOT}), a training-free sparsity allocation method.}
Treating each layer--step pair as a cell, ALLOT estimates each cell's contribution to final-video error through (i) cell probing and (ii) trajectory calibration, then jointly optimizes cell densities under a measured-latency budget. It preserves existing selection strategies and kernels, replacing only the allocation component.
At equal or shorter wall-clock time, ALLOT outperforms every compared allocation strategy with XAttention, with an 11.8\,dB PSNR gain on HunyuanVideo. Extensive experiments across four additional published pipelines on Wan2.1, Wan2.2, and HunyuanVideo show that ALLOT outperforms or ties the default in all nine settings.

\end{abstract}

\section{Introduction}
\label{sec:intro}

Diffusion Transformers (DiTs)~\cite{dit} have been widely adopted for video generation. Models such as Wan~\cite{wan} and HunyuanVideo~\cite{hunyuanvideo} produce high-quality, spatio-temporally consistent videos, yet the cost of their full 3D attention grows quadratically with the number of tokens~\cite{vaswani2017attention}.

On a single GPU, Wan2.1-14B takes roughly one hour to generate a 720p, 81-frame video, and attention accounts for more than half of the end-to-end time.

Sparse attention has been widely studied as a means of reducing this computational cost, with existing approaches broadly categorized as either training-free or training-based.
Many training-free methods~\cite{svg2,spargeattn,xattention,svoo} adopt \emph{block-sparse attention}, which partitions query and key tokens into blocks. These methods estimate the importance of each query--key block pair and compute dense attention only for selected pairs. {These works primarily address two questions: which blocks participate in the attention computation, and how to execute the resulting sparse computation efficiently at the kernel level.
}

{Notably, beyond determining which blocks participate in the computation, the number of blocks retained in each layer at each denoising step also significantly affects generation quality and latency.
In threshold-based methods~\cite{xattention,spargeattn,svoo}, the thresholds serve as an implicit allocation rule. 
With all other conditions identical, changing only the allocation can substantially alter final-video fidelity. 
Despite the existence of some explicit allocation approaches, existing methods generally have the following two limitations:}

\begin{itemize}[leftmargin=*]
  \item \textbf{Heuristic rules do not transfer reliably across settings.}
  Many works use fixed rules, such as keeping the first layer or the first 10--30\% of denoising steps dense~\cite{svoo,radial,svg2}. Our measurements in \S\ref{sec:motivation} show that effective allocations vary across models and selection strategies. Consequently, a rule tuned for one setting does not reliably transfer to another and must be re-tuned.

  \item \textbf{Local metrics do not capture final-video effects.}
  Automatic allocation methods commonly rely on local proxies, such as cumulative attention mass~\cite{xattention,svoo}, Hellinger distance~\cite{scalingattention}, and one-step output error~\cite{efficientvdit,heart}, rather than direct measurements of final-video error. Our analysis in \S\ref{sec:motivation} shows that the rankings induced by these proxies disagree with those based on final-video error, and allocations guided by these metrics can even yield lower final-video fidelity than a uniform baseline.
\end{itemize}

To address these limitations, we term each (layer, step) pair a \emph{cell} and evaluate it directly by its \emph{sensitivity}, defined as the degradation it induces in the final video relative to the dense output. Building on this formulation, we propose \textbf{A}llocating \textbf{L}ayer--step \textbf{L}atency by \textbf{O}utput \textbf{T}ests (ALLOT), a training-free latency allocation method.

ALLOT estimates each cell's sensitivity from two factors: cell probing measures the error injected at its step, and trajectory calibration measures how subsequent denoising steps amplify or attenuate that error.

It then jointly allocates all cell densities under a measured-latency budget, so warm-up length and dense layers emerge from the solution rather than hand-set rules.
Because sensitivity rankings remain stable across inputs for a fixed model and selection strategy (\S\ref{sec:motivation}), we perform both measurements once offline and reuse their results across prompts.
During inference, ALLOT looks up densities in a precomputed table and changes only the allocation in existing block-sparse attention pipelines.

Overall, our main contributions are threefold:

\begin{itemize}
  \item We show that final-video sensitivity varies by orders of magnitude across layers and steps and is misjudged by local metrics, while sensitivity rankings remain stable across inputs.
  \item We propose ALLOT, which estimates each cell's error by combining a one-step cell probe with a trajectory gain and jointly solves for all cell densities under a measured-latency budget, replacing only the allocation of existing pipelines.
  \item Extensive experiments across six selection strategies and three model families show that ALLOT outperforms or ties every compared allocation at equal or shorter wall-clock time~(\S\ref{sec:exp}).
\end{itemize}

\section{Related Work}
\label{sec:related}

\paragraph{Block-sparse attention.}
Block-sparse attention methods for video DiTs are either trained or training-free.
Trained methods (VSA~\cite{vsa}, VMoBA~\cite{vmoba}, NABLA~\cite{nabla}) train or fine-tune the model with sparse attention.
Training-free methods either apply static masks~\cite{sta,radial,scalingattention} or select blocks at run time~\cite{svg,spargeattn,svg2,xattention,svoo,adaspa}.
ALLOT retains the selection strategies and kernels of existing methods, replacing only the allocation~(\S\ref{sec:plugin}).

\paragraph{Allocation in existing methods.}
Existing methods allocate density, a threshold or a sparse pattern per layer~\cite{efficientvdit}, per (layer, head)~\cite{svoo,heart,sparsevdit}, per step~\cite{pasa,dfsattn}, per (layer, step)~\cite{ditfastattn}, or per (step, layer, head)~\cite{scalingattention,calibatt}, based on attention mass, Hellinger distance, one-step output error or other local metrics. Across steps, most use a heuristic warm-up or a preset shape, or treat all steps equally.
ALLOT also allocates per (layer, step) but calibrates each cell against the final video to capture cross-step error amplification and solves for all cells jointly under a measured-latency budget.

\section{Motivation and Key Observations}
\label{sec:motivation}

\begin{figure*}[t]
\centering
\includegraphics[width=\textwidth]{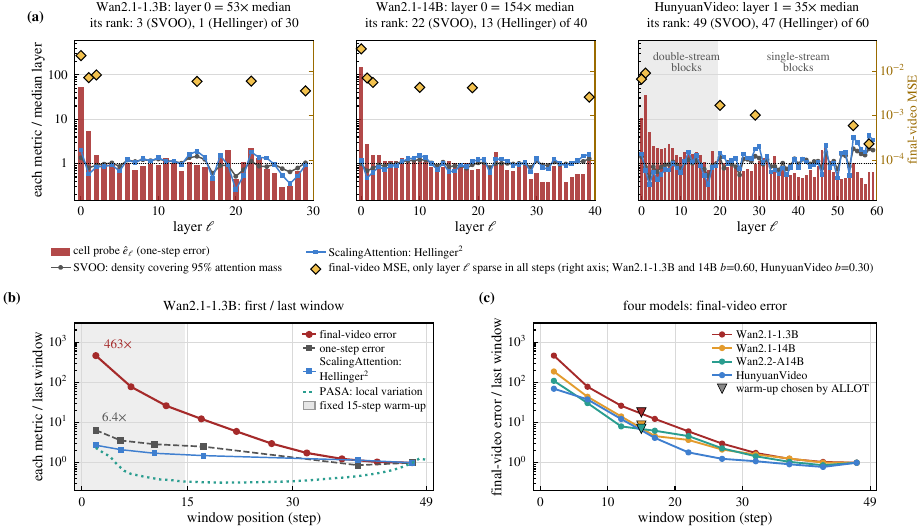}
\caption{\textbf{Layer and step sensitivity.}
(a)~Single-layer one-step error ($\dens{=}0.30$, $t{=}0.85$; bars), attention-map metrics~\cite{svoo,scalingattention} (lines), and final-video MSE with that layer sparse throughout (diamonds). Titles give the most sensitive layer and its metric ranks.
(b)~Wan2.1-1.3B window errors versus local metrics~\cite{pasa}; annotations give first/last-window ratios.
(c)~Window errors across four models.}
\label{fig:obs}
\end{figure*}

\paragraph{Setup.}
Let $\lay$ index layers and $s$ index denoising steps. The sensitivity $S(\lay,s;\dens)$ of cell $(\lay,s)$ measures how much sparsifying that cell to density~$\dens$ degrades the final video relative to the dense reference.
{Measuring this directly for every cell would require thousands of full generations and is therefore impractical.} We instead decompose it into two cheaper signals: the one-step output error when one layer is sparsified and the final-video error when a window of steps is sparsified (\S\ref{sec:probing}, \S\ref{sec:calibration}).

\paragraph{Observation 1: Layer sensitivity varies widely and is not reliably reflected by attention-map metrics.}
As shown in \cref{fig:obs}(a), the first one or two layers have one-step errors up to 154 times the median-layer error. When evaluated on the final video, the most sensitive layers coincide with these high-error layers.
Attention-map metrics (attention mass~\cite{svoo} and Hellinger distance~\cite{scalingattention}) yield different sensitivity rankings, and allocations based on them degrade fidelity (Appendix~\ref{app:B}).
Cell probes therefore provide a more reliable basis for layer allocation.
The most sensitive layer varies across models, and a fixed ``first layer dense'' rule therefore does not transfer reliably across settings.

\paragraph{Observation 2: Final-video effects vary sharply across denoising steps and are understated by one-step errors.}
As shown in \cref{fig:obs}(b, c), earlier windows produce markedly larger final-video errors than later ones under the same sparsification configuration. The first-to-last-window MSE ratio reaches several hundred for some models.
By contrast, the corresponding one-step errors show a much smaller difference and even reverse the ordering of the first and last windows on Wan2.2.
One-step errors therefore require trajectory calibration before guiding per-step density.

\paragraph{Observation 3: Sensitivity rankings remain stable across inputs for a fixed model and selection strategy.}
Across all tested models, per-input sensitivity rankings closely match rankings aggregated over calibration inputs, with median Spearman correlations of 0.80--1.00 (Appendix~\ref{app:D}).
Prior work reports similar input stability for attention-map metrics~\cite{svoo,scalingattention}.
Our results extend this observation to output-error sensitivity, allowing cell probes and trajectory gains to be measured offline and reused across prompts.
\section{ALLOT}
\label{sec:method}

\begin{figure*}[t]
\centering
\includegraphics[width=\textwidth]{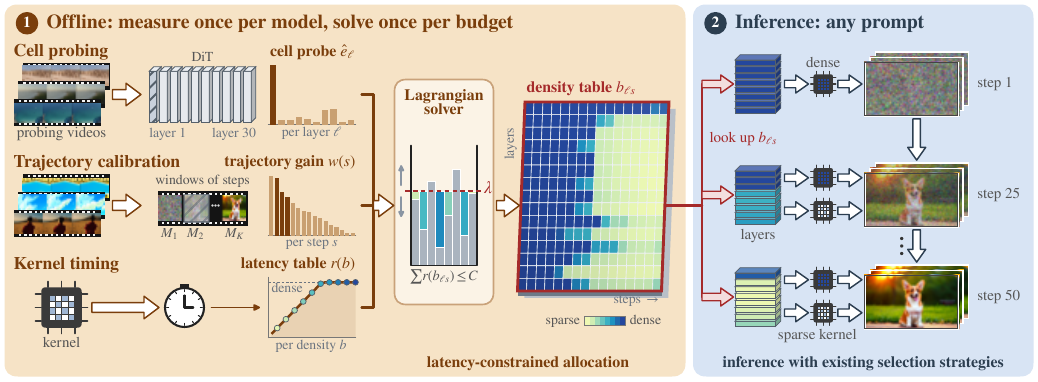}
\caption{\textbf{ALLOT overview.}
\textcircled{\scriptsize 1}~Offline: measure cell probes $\eprobe_\lay$, trajectory gains $w(s)$ and kernel latency $\lut(\dens)$; solve a density table for each latency budget~$C$ (\S\ref{sec:allocation}).
\textcircled{\scriptsize 2}~Inference: look up each cell's density for any prompt, using dense attention when $\dens_{\lay s}{=}1$ and the existing sparse kernel otherwise.}
\label{fig:method}
\end{figure*}

ALLOT jointly optimizes the density of every (layer, step) cell under a measured-latency budget, minimizing the estimated total sensitivity.
It consists of three modules: cell probing (\S\ref{sec:probing}), trajectory calibration (\S\ref{sec:calibration}) and latency-constrained allocation (\S\ref{sec:allocation}), built on the first-order analysis and calibrated estimate of \S\ref{sec:problem}. \S\ref{sec:plugin} describes inference and integration with existing pipelines.

\subsection{Problem formulation}
\label{sec:problem}

With thousands of cells in one denoising process, measuring each one on the final video is prohibitively expensive. We therefore approximate the final-video error as a sum of per-cell contributions and let a solver optimize the result.

\paragraph{First-order motivation and calibrated estimate.}
The error introduced at one step is carried to the final video by every subsequent step~\cite{epacache,lemica}.
Let $\epsilon_s$ denote the error that step~$s$'s sparse approximation introduces in the velocity prediction, and $\epsilon_{\lay s}$ denote the velocity-prediction change when only layer~$\lay$ is sparsified.
A first-order perturbation analysis around the dense trajectory approximates the final-latent deviation as
\begin{equation}
  \delta\approx\sum_sP_s\epsilon_s,
  \label{eq:deviation}
\end{equation}
where $P_s$ is the linearized propagation operator from a velocity error at step~$s$ to its effect on the final latent (Appendix~\ref{app:E}).
We make two approximations.
\textbf{(a)~Additivity}: layer perturbations superpose to first order within one step ($\epsilon_s\approx\sum_\lay\epsilon_{\lay s}$), and errors of different cells are orthogonal in expectation.
\textbf{(b)~Separability}: downstream amplification after step~$s$ depends only on~$s$, not on the layer (Appendix~\ref{app:E}).
Under these approximations,
\begin{equation}
  \mathbb{E}\lVert\delta\rVert^2\approx\sum_{\lay,s}g(s)\,\mathbb{E}\lVert\epsilon_{\lay s}\rVert^2,
  \label{eq:additive}
\end{equation}
where $g(s)$ is the propagation gain after step~$s$ in the first-order model.

Equation~\eqref{eq:additive} motivates factorizing each cell's contribution into within-step error and downstream amplification.
Because scalar within-step MSE does not specify how an error propagates, the analysis does not determine $w(s)$ (Appendix~\ref{app:E}).
We calibrate $w(s)$ using final videos generated with one window sparsified (\S\ref{sec:calibration}).
Together with the cell probes, this gives the estimate of the sensitivity defined in \S\ref{sec:motivation}:
\begin{equation}
  \hat S(\lay,s;\dens)=w(s)\,\eprobe_\lay(t_s,\dens).
  \label{eq:estimate}
\end{equation}
Here $t_s$ is the step's noise level and $\eprobe$ is the within-step cell probe (\S\ref{sec:probing}).
The two factors serve complementary roles: the cell probe distinguishes layers and candidate densities within a step, whereas $w(s)$ places different steps on a common final-video-error scale.
Only relative sensitivities affect allocation: uniformly scaling the estimated error does not change the optimal table at a fixed latency budget.

\paragraph{Optimization problem.}
Given a latency budget $C$, find the density $\dens_{\lay s}$ of every cell:
\begin{equation}
  \min_{\dens_{\lay s}}\;\sum_{\lay,s}\hat S(\lay,s;\dens_{\lay s})\quad
  \text{s.t.}\;\sum_{\lay,s}\lut(\dens_{\lay s})\le C,
  \label{eq:opt}
\end{equation}
where $\dens_{\lay s}{=}1$ means fully dense (error counted as zero).
$\lut(\dens)$ is the measured wall-clock time of a single cell's attention at density~$\dens$ (the latency table, \S\ref{sec:allocation}). $\lut$ and $C$ are both normalized so that a dense cell equals~1.
Because non-attention computation does not change with allocation, constraining attention time approximately constrains end-to-end latency.

\subsection{Cell probing on the model output}
\label{sec:probing}

The cell probe $\eprobe_\lay(t,\dens)$ measures a layer's output error within one step. It is used to compare layers within the same step (Observation~1) and to compare one cell's error across densities.
We add noise at level~$t$ to a small number of real videos, sparsify only layer~$\lay$ to density~$\dens$ while keeping all other layers dense, and measure the MSE between this step's velocity prediction and the dense model's prediction~\cite{efficientvdit}.
At deployment, all other layers are sparse at the same time, unlike the measurement condition. This difference is analyzed in Appendix~\ref{app:F}.

Each measurement needs only one forward pass.
By Observation~3, layer rankings are input-stable, so a few probing videos suffice (\S\ref{sec:exp:robust}). Since rankings differ across selection strategies, probes are by default measured under the selection strategy used at deployment.
We tabulate probes on a grid over density and noise level, average over the probing videos, and interpolate in $\log\dens$--$\log\eprobe$ when solving (grid in \S\ref{sec:exp:setup}).

\subsection{Trajectory calibration against the final video}
\label{sec:calibration}

The trajectory gain $w(s)$ supplies the inter-step amplification that one-step errors cannot capture (Observation~2).
It reuses the window experiments of \cref{fig:obs}(b, c). Denoising steps are divided into contiguous windows. Only window~$\Omega_k$ has all layers sparsified to a fixed window density~$\dens_0$, with the remaining steps dense. The generated video's pixel MSE against the dense reference is~$M_k$.
Let $E_k=\sum_{s\in\Omega_k}\sum_\lay\eprobe_\lay(t_s,\dens_0)$ be the sum of the probes of all cells sparsified in this run. The gain is the final-video error per unit of probe error, shared by all steps in the window:
\begin{equation}
  w(s)=M_k/E_k,\qquad s\in\Omega_k.
  \label{eq:trajgain}
\end{equation}
By construction, summing Equation~\eqref{eq:estimate} over the window's cells recovers the measured $M_k$. The solver uses only the product $w(s)\,\eprobe_\lay(t_s,\dens)$.
Calibrating on sparse-window runs retains the error structure of the sparse-attention path used to measure $M_k$.
The resulting $w(s)$ captures downstream propagation, average cross-layer interactions and latent-to-pixel conversion. Appendix~\ref{app:E} provides an oracle diagnostic of this choice.
Sharing $w(s)$ across layers is the separability approximation.
Each window therefore needs one generation per prompt with all layers sparsified together, independent of the number of layers and the density grid (settings in \S\ref{sec:exp:setup}).

\subsection{Latency-constrained allocation}
\label{sec:allocation}

\paragraph{Latency table.}
$\lut(\dens)$ is the time of one attention call at density~$\dens$ divided by that of dense FlashAttention~\cite{flashattention2} on the same shape, measured with the kernel and resolution used at deployment.
$\lut(\dens)$ is not proportional to density. A sparse kernel is faster than dense only below a crossover density. Assuming cost proportional to density or FLOPs would therefore overestimate the gain from sparsity.
When solving, $\lut$ is clipped to~1, and cells whose density would lie above the crossover are set to dense.

\paragraph{Solver.}
Equation~\eqref{eq:opt} is solved by Lagrangian relaxation~\cite{efficientvdit}.
The candidate density set~$\mathcal{B}$ is a set of equally spaced sparse densities plus dense ($\dens{=}1$). Cell probes are interpolated to~$\mathcal{B}$ and the step's noise level, then multiplied by~$w(s)$ to obtain $\hat S(\lay,s;\dens)$ for each candidate.
The solver runs a bisection on~$\lambda$. Given~$\lambda$, each cell independently picks the density minimizing $\hat S(\lay,s;\dens)+\lambda\,\lut(\dens)$ over~$\mathcal{B}$, yielding a density table. If the table's total cost exceeds~$C$, $\lambda$ is increased. Otherwise, it is decreased. This process continues until convergence (Algorithm~1, Appendix~\ref{app:F}).

\begin{proposition}
\label{prop:lagrangian}
For any $\lambda\ge0$, the density table $\dens^\lambda$ obtained by independently minimizing $\hat S(\lay,s;\dens)+\lambda\,\lut(\dens)$ over~$\mathcal{B}$ for each cell is an exact optimum of Eq.~\eqref{eq:opt} (restricted to~$\mathcal{B}$) when the budget equals the table's own total cost~\cite{everett}. The result applies to the discrete estimated objective and requires no convexity (Appendix~\ref{app:E}).
\end{proposition}
The bisection locates the two Lagrangian tables bracketing~$C$, which typically differ in only one cell (Appendix~\ref{app:E}).
Warm-up length, which layers stay dense and how density varies across steps all emerge from the solver without heuristic rules (Appendices~\ref{app:F} and~\ref{app:J}).

\subsection{Inference and integration with existing selection strategies}
\label{sec:plugin}

\paragraph{Inference.}
The density table is solved offline, once per model, selection strategy and latency budget.
At generation time, the density of each layer is set from the table before each step~$s$. Cells with $\dens_{\lay s}{=}1$ use the original dense attention. All other cells are sparsified by the selection strategy at density~$\dens_{\lay s}$ and computed by the original kernel.
The same table is used for all prompts. Changing the budget only requires re-solving one table.

\paragraph{Integration.}
ALLOT keeps each selection strategy's kernel and ranking, replacing only the cutoff that determines how many ranked blocks or clusters are retained.
Threshold-based defaults (SVOO, SVG2, SpargeAttn and XAttention) are converted to a density cutoff. For SpargeAttn, this uses its official top-$k$ API.
This removes native per-head or per-input adaptation. Splitting by SVOO's head densities is 1.44\,dB worse than an equal split (Appendices~\ref{app:B} and~\ref{app:F}).
Throughout, ``changing only the allocation'' refers to this integration.

\paragraph{Supported interfaces.}
ALLOT applies when a selection strategy supports a controllable per-cell density and its latency can be measured. Fixed-mask strategies such as Radial are restricted to densities reachable by their mask parameters.
Density denotes retained blocks for mean-pooled top-$k$, SpargeAttn, Radial and XAttention, and retained key tokens for the variable-size clusters of SVOO and SVG2.

\section{Experiments}
\label{sec:exp}

\subsection{Experimental setup}
\label{sec:exp:setup}

\paragraph{Models and data.}
Four models: Wan2.1-T2V-1.3B and 14B (hereafter 1.3B and 14B), Wan2.2-T2V-A14B~\cite{wan} and HunyuanVideo~\cite{hunyuanvideo}.
Resolution and frame count are given in each table, following a published setting where one exists. The analyses in \S\ref{sec:motivation} use 480p$\times$33. All use 50 denoising steps.
Wan2.2 results are in \cref{tab:crossmodel,tab:plugin}, and HunyuanVideo results are in \cref{tab:plugin} and Appendices~\ref{app:G} and~\ref{app:J}.
The evaluation set consists of 32 VBench~\cite{vbench} prompts.

\paragraph{Metrics.}
Fidelity metrics (PSNR, SSIM~\cite{ssim} and LPIPS~\cite{lpips}) measure similarity to the dense reference (same prompt and seed) rather than absolute generation quality.
Perceptual quality is measured by VBench~\cite{vbench}, weighted over the six dimensions used by HEART~\cite{heart}.
At matched or shorter wall-clock time, a win requires a positive lower bound of the paired 95\% PSNR-difference interval and lower mean LPIPS, while a loss requires a negative upper bound and higher mean LPIPS.
All other cases are ties, and VBench does not enter the verdict (Appendix~\ref{app:A}).

\paragraph{Baselines.}
Published pipelines~\cite{xattention,svoo,spargeattn,radial,svg2} follow their documented settings (SVG2 uses HEART's five-step warm-up), while allocation baselines~\cite{heart,efficientvdit,scalingattention,pasa,calibatt,dfsattn,ditfastattn} are re-implemented from their papers (Appendices~\ref{app:G} and~\ref{app:H}).
For mean-pooled top-$k$, the uniform baseline keeps the first 15 steps dense. The strengthened baseline additionally keeps layer~0 dense and tunes the warm-up on 1.3B (Appendix~\ref{app:J}).

\paragraph{Implementation details.}
Mean-pooled top-$k$ (the coarse stage of VSA~\cite{vsa}) reorders tokens cube by cube ($3{\times}4{\times}4$ cubes) and groups every 64 consecutive tokens into a block.
Cell probes are measured on four OpenVid~\cite{openvid} real videos over a grid of 8 densities $\times$ 9 noise levels.
Trajectory gains are measured using 10 five-step windows at window density $\dens_0{=}0.30$. 

\subsection{Comparison with other allocation methods}
\label{sec:exp:compare}

To compare allocation strategies in isolation, we fix XAttention's selection strategy and kernel, changing only the allocation (\cref{tab:compare}).
ALLOT outperforms every compared allocation method, all of which use hand-set rules or local metrics, by 1.53--7.31\,dB.

The effect of density truncation depends on the selection strategy: it lowers SVOO's PSNR but improves XAttention's.
ALLOT further improves on XAttention's density-cutoff default at matched wall-clock time (Appendix~\ref{app:F}).

VBench scores remain close to those of the baselines.
ALLOT also outperforms every compared allocation method on additional XAttention settings spanning higher resolution and HunyuanVideo, all at equal or shorter wall-clock time (Appendix~\ref{app:G}).

\begin{table*}[t]
\centering
\small
\caption{\textbf{Comparison with other allocation methods} (XAttention on Wan2.1-1.3B 480p$\times$81, $n{=}32$).
Default: XAttention's published configuration. Rows grouped by wall-clock time, with ALLOT re-solved for each of the first two groups. The third group applies HEART's cross-step mask reuse (TMR) to both sides. Bold: best PSNR/SSIM/LPIPS in group. Details in Appendix~\ref{app:G}.}
\label{tab:compare}
\begin{tabular*}{\textwidth}{@{\extracolsep{\fill}}lccccrr@{}}
\toprule
Allocation & PSNR$\uparrow$ & SSIM$\uparrow$ & LPIPS$\downarrow$ & VBench$\uparrow$ & Wall-clock (s) & Speedup \\
\midrule
Dense & --- & --- & --- & 72.35 & 183.8 & 1.00$\times$ \\
Default & 18.14 & 0.659 & 0.268 & 70.27 & 148.7 & 1.24$\times$ \\
CalibAtt-style & 19.67 & 0.700 & 0.222 & 70.79 & 149.6 & 1.23$\times$ \\
ALLOT & \textbf{25.37} & \textbf{0.835} & \textbf{0.106} & 71.24 & 148.6 & 1.24$\times$ \\
\addlinespace
HEART (EBC) & 19.43 & 0.699 & 0.221 & 72.02 & 151.7 & 1.21$\times$ \\
Efficient-vDiT & 19.08 & 0.702 & 0.225 & 72.32 & 151.1 & 1.22$\times$ \\
DiTFastAttn-style & 19.38 & 0.708 & 0.219 & 72.27 & 152.2 & 1.21$\times$ \\
ScalingAttention (FAST) & 22.75 & 0.790 & 0.141 & 71.09 & 152.9 & 1.20$\times$ \\
PASA & 21.91 & 0.768 & 0.163 & 71.25 & 151.8 & 1.21$\times$ \\
DFSAttn-style & 24.86 & 0.834 & 0.107 & 71.20 & 151.3 & 1.21$\times$ \\
ALLOT & \textbf{26.39} & \textbf{0.854} & \textbf{0.089} & 71.38 & 151.3 & 1.21$\times$ \\
\addlinespace
HEART (EBC + TMR) & 19.54 & 0.706 & 0.214 & 71.62 & 145.0 & 1.27$\times$ \\
ALLOT + TMR & \textbf{26.13} & \textbf{0.847} & \textbf{0.094} & 71.34 & 146.0 & 1.26$\times$ \\
\bottomrule
\end{tabular*}
\end{table*}

\begin{figure*}[t]
\centering
\includegraphics[width=\textwidth,height=0.38\textheight,keepaspectratio]{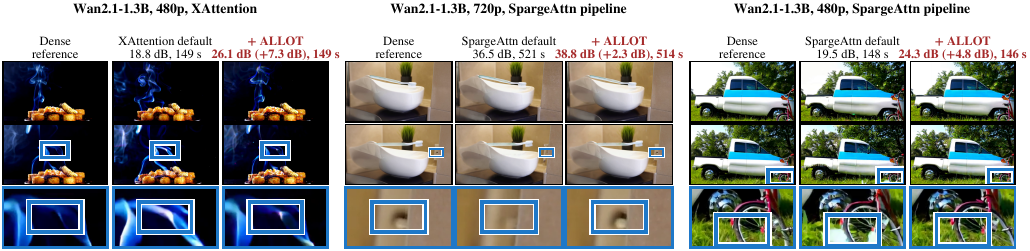}
\caption{\textbf{Qualitative comparison.} Dense, baseline and +\,ALLOT (left to right) at the first and last frames; bottom rows enlarge the boxed regions, with inner boxes marking larger baseline deviations from dense. Headers give PSNR to dense, gain and wall-clock time. Prompts are near the median gain. Settings: \cref{tab:compare} (XAttention), Appendix~\ref{app:H} (SpargeAttn 720p), and \cref{tab:plugin} (SpargeAttn 480p).}
\label{fig:qual}
\end{figure*}

\subsection{Integration with published pipelines}
\label{sec:exp:plugin}

We integrate ALLOT into the official pipelines of SVOO~\cite{svoo}, SVG2~\cite{svg2}, SpargeAttn~\cite{spargeattn} and Radial~\cite{radial}, changing only the allocation (\cref{tab:plugin}, details in Appendix~\ref{app:H}).
ALLOT outperforms the default in six of the nine settings (by up to 5.03\,dB in PSNR) and ties in the other three (Appendix~\ref{app:H}).
Notably, on HunyuanVideo (13B) with SVOO, ALLOT improves PSNR by 3.38\,dB while increasing end-to-end speedup from 1.75$\times$ to 1.90$\times$.
\cref{fig:qual} shows qualitative comparisons.

\begin{figure}[t]
\centering
\includegraphics[width=\columnwidth]{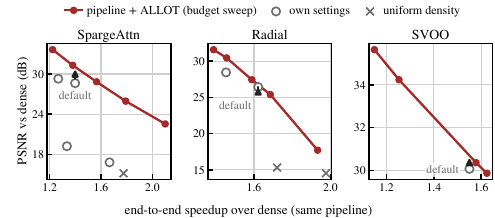}
\caption{\textbf{Fidelity--speedup curves} (Wan2.1-1.3B 720p; prompts as in \cref{tab:plugin}; y-ranges differ). Open circles: pipeline settings (SpargeAttn: default, two released and one retuned checkpoint; Radial: default and decay 1.0; SVOO: default). Arrows show gains at the default's speed.}
\label{fig:pareto}
\end{figure}

To compare across speeds, we solve ALLOT's tables at several latency budgets on three published pipelines (1.3B 720p) and plot PSNR against end-to-end speedup (\cref{fig:pareto}).
All of the pipelines' own settings (open circles) lie below ALLOT's curve.

\begin{table*}[t]
\centering
\small
\caption{\textbf{Integration with published pipelines, changing only the allocation} ($n{=}32$; speedup = dense wall-clock / row wall-clock; dense times and paired intervals in Appendix~\ref{app:H}).
Bold: best PSNR/SSIM/LPIPS in each setting.}
\label{tab:plugin}
\begin{tabular*}{\textwidth}{@{\extracolsep{\fill}}llccccrr@{}}
\toprule
Pipeline \& setup & Allocation & PSNR$\uparrow$ & SSIM$\uparrow$ & LPIPS$\downarrow$ & VBench$\uparrow$ & Wall-clock (s) & Speedup \\
\midrule
Wan2.1-1.3B 480p$\times$81 & Default & 22.67 & 0.768 & 0.150 & 70.03 & 147.7 & 1.24$\times$ \\
SpargeAttn & +\,ALLOT & \textbf{27.70} & \textbf{0.876} & \textbf{0.070} & 71.58 & 145.8 & 1.26$\times$ \\
\midrule
Wan2.1-1.3B 720p$\times$81 & Default & 26.44 & 0.832 & 0.143 & 71.71 & 448.9 & 1.62$\times$ \\
Radial & +\,ALLOT & \textbf{26.83} & \textbf{0.837} & \textbf{0.138} & 71.08 & 441.5 & 1.65$\times$ \\
\midrule
Wan2.1-1.3B 720p$\times$81 & Default & 24.25 & 0.808 & \textbf{0.151} & 71.09 & 410.9 & 1.66$\times$ \\
SVG2 & +\,ALLOT & \textbf{25.60} & \textbf{0.819} & 0.156 & 70.75 & 410.5 & 1.66$\times$ \\
\midrule
Wan2.1-1.3B 720p$\times$81 & Default & 30.08 & \textbf{0.893} & \textbf{0.082} & 72.12 & 451.6 & 1.55$\times$ \\
SVOO & +\,ALLOT & \textbf{30.73} & 0.890 & 0.095 & 71.45 & 447.7 & 1.56$\times$ \\
\midrule
Wan2.1-14B 720p$\times$81 & Default & 24.75 & 0.839 & 0.128 & 71.70 & 2296.9 & 1.56$\times$ \\
Radial & +\,ALLOT & \textbf{25.28} & \textbf{0.850} & \textbf{0.115} & 71.65 & 2290.7 & 1.56$\times$ \\
\midrule
Wan2.1-14B 720p$\times$81 & Default & 28.15 & 0.903 & 0.068 & 71.67 & 2603.5 & 1.31$\times$ \\
SVOO & +\,ALLOT & \textbf{31.70} & \textbf{0.930} & \textbf{0.044} & 71.59 & 2586.6 & 1.32$\times$ \\
\midrule
Wan2.2-A14B 720p$\times$81 & Default & 25.36 & 0.853 & 0.089 & 74.25 & 2735.8 & 1.23$\times$ \\
SVOO & +\,ALLOT & \textbf{30.23} & \textbf{0.911} & \textbf{0.044} & 73.93 & 2732.2 & 1.23$\times$ \\
\midrule
HunyuanVideo 720p$\times$129 & Default & \textbf{28.08} & \textbf{0.893} & \textbf{0.112} & 70.96 & 1814.2 & 1.76$\times$ \\
Radial & +\,ALLOT & 27.53 & 0.892 & 0.121 & 70.55 & 1805.2 & 1.77$\times$ \\
\midrule
HunyuanVideo 720p$\times$129 & Default & 27.55 & 0.884 & 0.118 & 71.71 & 1842.7 & 1.75$\times$ \\
SVOO & +\,ALLOT & \textbf{30.93} & \textbf{0.923} & \textbf{0.071} & 71.46 & 1697.0 & 1.90$\times$ \\
\bottomrule
\end{tabular*}
\end{table*}

\subsection{Cross-model comparison}
\label{sec:exp:crossmodel}

On the three Wan models with mean-pooled top-$k$, ALLOT outperforms both the uniform and the strengthened baseline (\cref{tab:crossmodel}). We extend this comparison to HunyuanVideo at equal density in Appendix~\ref{app:J}.
\cref{fig:alloc} shows what a fixed rule misses: the first few layers and certain deeper layers stay dense longest, while density decreases over several intermediate steps rather than dropping at a single step.

The best rule also changes with the budget, so we test the baselines at other budgets: ALLOT remains above the uniform baseline across budgets at 720p$\times$81 (1.3B and 14B, Appendix~\ref{app:H}). On 1.3B 480p$\times$33, it also outperforms the strengthened baseline with its warm-up re-searched at each of three budgets (Appendix~\ref{app:J}).

\begin{table*}[t]
\centering
\small
\caption{\textbf{Cross-model comparison} (mean-pooled top-$k$ selection with the VSA kernel; groups differ only in density tables; $\Delta$PSNR is the mean paired difference). Details in Appendix~\ref{app:J}.}
\label{tab:crossmodel}
\begin{tabular*}{\textwidth}{@{\extracolsep{\fill}}lccccc@{}}
\toprule
Model \& setup & Speedup & $n$ & PSNR$\uparrow$ (uniform $\to$ ALLOT) & $\Delta$PSNR vs.\ uniform & $\Delta$PSNR vs.\ strengthened \\
\midrule
Wan2.1-1.3B, 480p$\times$33 & 1.14$\times$ & 32 & 19.59 $\to$ 25.99 & $+$6.40 & $+$2.55 \\
Wan2.1-14B, 720p$\times$81 & 1.29$\times$ & 16 & 24.87 $\to$ 31.25 & $+$6.38 & $+$0.76 \\
Wan2.2-A14B, 480p$\times$33 & 1.09$\times$ & 32 & 19.36 $\to$ 25.86 & $+$6.50 & $+$1.67 \\
\bottomrule
\end{tabular*}
\end{table*}
\begin{figure}[t]
\centering
\includegraphics[width=\columnwidth]{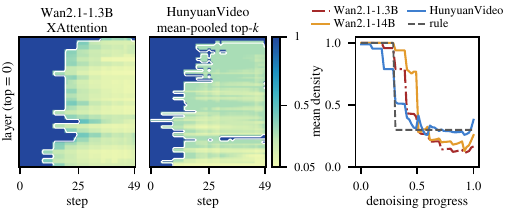}
\caption{\textbf{Solved density tables} for XAttention (\cref{tab:compare}) and HunyuanVideo (equal density; Appendix~\ref{app:J}); white contours mark dense cells. Right: mean density per step, including Wan2.1-14B (\cref{tab:crossmodel}), versus a rule keeping the first 30\% of steps dense and using 0.30 thereafter.}
\label{fig:alloc}
\end{figure}

\subsection{Ablation study}
\label{sec:exp:ablation}

Each ablation replaces one component in the XAttention setup of \cref{tab:compare} at ALLOT's cost.
Trajectory calibration is dominant: setting $w(s)=1$ for every step lowers PSNR by 7.05\,dB, and DiTFastAttn's greedy rule on the same one-step errors reaches a similar 19.38\,dB (\cref{tab:compare,tab:ablation}).
A shared density across layers and a parametric shape rule also underperform, whereas density-proportional cost makes no significant difference after wall-clock alignment.
Mean-pooled top-$k$ ablations and an LPIPS-calibrated variant are reported in Appendix~\ref{app:I}.

\subsection{Verification, robustness and offline cost}
\label{sec:exp:robust}

\paragraph{Verifying the sensitivity estimate.}
The solver ranks density tables by their estimated total sensitivity, the sum of $\hat S$ over all cells (Eq.~\eqref{eq:estimate}).
We validate this ranking against measured final-video error.
At fixed total cost, permuting ALLOT's densities among cells lowers PSNR on every prompt, and the measured ordering matches the prediction.
Across 80 tables at multiple budgets on held-out prompts, the predicted and measured rankings agree strongly (pooled Kendall $\tau=0.83$, Appendix~\ref{app:E}).

\paragraph{Number of calibration inputs.}
Cell probes use four probing videos and trajectory calibration uses eight prompts.
With a single probing video or calibration prompt, most cells keep the same dense/sparse status.
On the XAttention setting of \cref{tab:compare}, eight probing videos bring no significant gain over four, and one or two lose less than a tenth of ALLOT's gain (Appendix~\ref{app:K}).
Window density and probe resolution have mild effects. Calibration with SpargeAttn rather than mean-pooled top-$k$ gives a 0.993 correlation in $w$ and 96.0--96.5\% cell-decision agreement (Appendix~\ref{app:K}).
\paragraph{Offline cost.}
Measurement takes roughly 6--23 GPU-hours per model (over 30 for the two experts of Wan2.2) and is done once per model and selection strategy (Appendix~\ref{app:F}).
Changing the latency budget or the number of steps only requires re-solving the table, which takes about 4\,s on CPU. With 30 steps, CFG 7.5 or the Euler sampler, ALLOT still outperforms the default without re-measuring (Appendix~\ref{app:K}).

\begin{table}[h]
\centering
\small
\setlength{\tabcolsep}{3pt}
\caption{\textbf{Ablation} (XAttention, Wan2.1-1.3B 480p$\times$81, $n{=}32$; one module replaced per row at the same cost as ALLOT; full metrics in Appendix~\ref{app:I}).}
\label{tab:ablation}
\vspace{-6pt}
\begin{tabular}{@{}llcc@{}}
\toprule
Module & Replaced with & PSNR & $\Delta$PSNR \\
\midrule
--- & ALLOT & 26.39 & --- \\
Cell prob. & Same density for all layers & 25.32 & $-$1.07 \\
Traj.\ cal. & $w \equiv 1$ & 19.34 & $-$7.05 \\
Per-cell solve & Parametric shape rule & 24.63 & $-$1.76 \\
Latency table & Density-proportional cost & 26.27 & $-$0.12 \\
\bottomrule
\end{tabular}
\vspace{-8pt}
\end{table}

\section{Conclusion}
\label{sec:conclusion}

We showed that layer--step sensitivity varies across models and selection strategies, is not reliably captured by local metrics, yet is nearly input-invariant for a fixed model and strategy.
ALLOT turns these findings into a training-free allocator by calibrating one-step errors against final-video damage and optimizing a latency-constrained density table.
Ablations identify trajectory calibration as the dominant component, while the same offline measurements remain effective across budgets, step counts, guidance scales and samplers without re-measurement.
Across three model families, ALLOT outperforms every compared allocation method on XAttention and outperforms or ties the default on four other published pipelines, all at equal or shorter wall-clock time.

\paragraph{Limitations.}
ALLOT requires offline calibration, and switching selection strategies requires new cell probes.
Its additive, layer-separable estimate predicts rankings more reliably than error magnitudes, while the common density interface replaces native threshold adaptation.
Modeling cell interactions and preserving adaptive thresholds are directions for future work.
\vfill

{
    \small
    \bibliographystyle{ieeenat_fullname}
    \bibliography{main}
}

\providecommand{\dci}[3]{$#1$\,[$#2$,\,$#3$]}                 
\providecommand{\pd}[4]{\dci{#1}{#2}{#3}\ #4}                  
\providecommand{\suppfit}[1]{\resizebox{\ifdim\width>\linewidth\linewidth\else\width\fi}{!}{#1}}  
\providecommand{\hd}[1]{\begin{tabular}[b]{@{}c@{}}#1\end{tabular}}                   
\setcounter{dbltopnumber}{3}
\makeatletter
\providecommand{\suppsubsec}[3]{
  \subsection*{#1.~#2}%
  \phantomsection\def\@currentlabel{#1}\label{#3}}
\makeatother

\newcommand{\readablepairedtable}{%
  \footnotesize
  \setlength{\tabcolsep}{2pt}%
  \renewcommand{\arraystretch}{1.15}%
  \renewcommand{\pd}[4]{\begin{tabular}{@{}c@{}}$##1$\\$[##2,\,##3]$\\##4\end{tabular}}}

\maketitlesupplementary
\appendix
\renewcommand{\thesection}{\Alph{section}}
\counterwithin{table}{section}
\counterwithin{figure}{section}

\section{Statistical Conventions and Timing}
\label{app:A}

\subsection{Metrics and verdicts}
\label{app:A1}

Ratios of final-video error are ratios of pixel MSEs.

\paragraph{Verdicts.}
A \emph{win} requires the lower bound of the paired PSNR interval to be $>0$ and the mean LPIPS to be lower; a \emph{loss} requires the upper bound of the PSNR interval to be $<0$ and the mean LPIPS to be higher; everything else is a \emph{tie}.
The appendix tables additionally list DISTS~\cite{dists} as a supplement only.
Paired differences are ALLOT $-$ baseline, followed by the 95\% $t$-interval over prompts and the numbers of prompts on which ALLOT is better/worse.

\paragraph{Trajectory gain.}
The trajectory gain is calibrated once per model family (Wan2.1 on 1.3B 480p$\times$33 with mean-pooled top-$k$, Wan2.2-A14B, and HunyuanVideo 480p$\times$33), and its window experiments are shared by all tables of that family across selection strategies, resolutions and model sizes. SVG2's official pipeline uses a different flow shift, so its window experiments are rerun on that schedule.

\subsection{Timing and equal wall-clock time}
\label{app:A2}

Wall-clock time is the per-video median in seconds, including text encoding, denoising and VAE decoding. Speedups are relative to each pipeline's own dense time (the dense times are in the Dense row of Table~\ref{tab:compare} and in Appendix~\ref{app:H}).
ALLOT's wall-clock time is considered matched when it is no more than $1.01\times$ the baseline’s.

\section{Observation 1: Across Layers}
\label{app:B}

\subsection{Cell probes and attention-map metrics}
\label{app:B1}

With each layer sparsified alone to $\dens{=}0.30$ (mean-pooled top-$k$), we record the cell probe and two attention-map metrics: the density needed to cover 95\% of the attention mass (SVOO's definition) and Hellinger$^2$ (the square of ScalingAttention's metric).

\begin{table*}[t]
\centering\ttiny
\caption{Cell probes and attention-map metrics. Ratio: cell probe of the most sensitive layer $\div$ that of the median layer; ranks are in decreasing order.}
\label{tab:B1}
\suppfit{\begin{tabular}{@{}lccccccc@{}}
\toprule
Model & $t$ & \hd{Most sensitive\\layer} & Ratio & \hd{Rank:\\coverage density} & \hd{Rank:\\$H^2$} & \hd{Spearman\\(coverage, $\eprobe$)} & \hd{Spearman\\($H^2$, $\eprobe$)} \\
\midrule
Wan2.1-1.3B & 0.85 & 0 & 53$\times$ & 3/30 & 1/30 & $+0.33$ & $+0.43$ \\
Wan2.1-1.3B & 0.45 & 1 & 27$\times$ & 26/30 & 26/30 & $+0.29$ & $+0.45$ \\
Wan2.1-1.3B & 0.15 & 0 & 15$\times$ & 1/30 & 1/30 & $+0.33$ & $+0.39$ \\
Wan2.1-14B & 0.85 & 0 & 154$\times$ & 22/40 & 13/40 & $+0.03$ & $+0.08$ \\
Wan2.1-14B & 0.45 & 0 & 75$\times$ & 11/40 & 6/40 & $-0.16$ & $-0.02$ \\
Wan2.1-14B & 0.15 & 0 & 71$\times$ & 7/40 & 4/40 & $-0.15$ & $+0.11$ \\
HunyuanVideo & 0.85 & 1 & 35$\times$ & 49/60 & 47/60 & $-0.26$ & $-0.20$ \\
HunyuanVideo & 0.45 & 1 & 23$\times$ & 51/60 & 48/60 & $-0.44$ & $-0.36$ \\
HunyuanVideo & 0.15 & 1 & 30$\times$ & 48/60 & 42/60 & $-0.51$ & $-0.42$ \\
\bottomrule
\end{tabular}}
\end{table*}

\begin{figure*}[t]
\centering
\includegraphics[width=\textwidth]{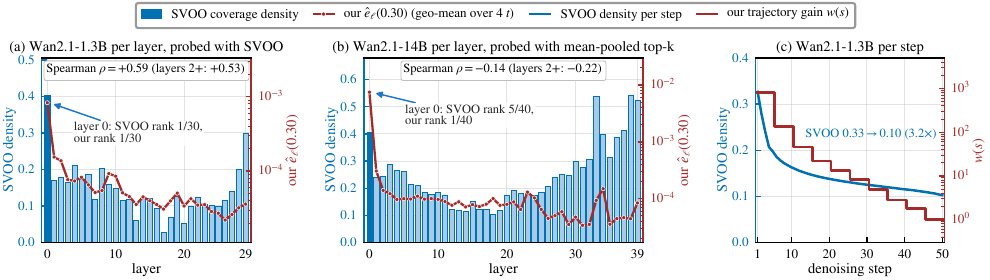}
\caption{SVOO's released per-layer coverage density (bars) and each layer's cell probe (line, $\dens{=}0.30$). (a)~Wan2.1-1.3B; (b)~Wan2.1-14B; (c)~step-wise trend: both decrease along denoising.}
\label{fig:B1}
\end{figure*}

The layer rankings of the attention-map metrics agree partly with that of the cell probe on 1.3B, are essentially unrelated to it on 14B, and are reversed on HunyuanVideo.
The magnitudes also differ sharply: on 1.3B the coverage density of layer~0 is only 1.36--1.84$\times$ that of the median layer, while its cell probe is 53$\times$, 8$\times$ and 15$\times$ at $t$ = 0.85, 0.45 and 0.15 (at $t$ = 0.45 the most sensitive layer is layer~1, at 27$\times$).
On 14B the coverage ratio is 0.99--1.17$\times$ and the cell-probe ratio is 154$\times$, 75$\times$ and 71$\times$.
On 14B the Spearman correlation between the cell probe and SVOO's released per-layer densities is $-0.14$ ($+0.59$ on 1.3B with probes under SVOO's selection strategy, Figure~\ref{fig:B1}).

\subsection{Allocating by other cross-layer metrics}
\label{app:B2}

\paragraph{\texorpdfstring{Hellinger$^2$}{Hellinger\texttwosuperior}.}
We replace only the solver's cross-layer signal, using Hellinger$^2$ instead of the cell probe while leaving all other inputs unchanged.
Because the solver optimizes all cells jointly, re-solving also changes the step allocation.
The paired PSNR differences in dB (ALLOT minus the Hellinger$^2$-guided allocation) are
1.3B 480p \pd{+3.62}{+3.10}{+4.15}{32/0},
14B 480p \pd{+5.88}{+5.13}{+6.64}{32/0}, and
HunyuanVideo 480p \pd{+1.38}{+1.07}{+1.68}{128/32}
($n{=}160$: the evaluation set plus 128 VBench all-dimension prompts).

\paragraph{SVOO's default density.}
We compare ALLOT with an allocation proportional to the actual per-cell densities produced by SVOO's default run (hereafter the \emph{default density}).
Both use the same dense cells and the same total budget at each step, differing only in how density is distributed across layers.
On HunyuanVideo 480p$\times$33, the paired PSNR difference in dB is
\pd{+2.40}{+1.93}{+2.87}{31/1};
LPIPS and DISTS also favor ALLOT (31/1 each).

\subsection{Per-head allocation within a cell}
\label{app:B3}

Each cell in the table takes a single density, and every head in that cell keeps the same fraction of its blocks.
On SVOO 1.3B 720p$\times$81, we compare this with a per-head split that keeps each cell's total but gives each head blocks in proportion to its density in SVOO's default run (more blocks for heads with dispersed attention, fewer for heads with concentrated attention).
ALLOT's equal split yields a PSNR 1.44\,dB higher than the per-head split (\pd{+1.44}{+1.05}{+1.84}{16/0}), and LPIPS and DISTS are also better (16/0 and 15/1; $n{=}16$).

\section{Observation 2: Across Steps}
\label{app:C}

\subsection{Window experiments}
\label{app:C1}

The 50 steps are split into 10 windows of 5 steps. Each run sparsifies all layers to 0.30 within one window only, keeps the other steps dense, and uses 8 VBench prompts per window.

\begin{table*}[t]
\centering\ttiny
\caption{Window experiments. Each entry: PSNR (in parentheses, final-video MSE relative to window 10, from 8-prompt mean).}
\label{tab:C1}
\suppfit{\begin{tabular}{@{}llcccc@{}}
\toprule
Window & Steps & Wan2.1-1.3B & Wan2.1-14B & Wan2.2-A14B & HunyuanVideo \\
\midrule
1 & 0--4 & 10.71 (463) & 10.70 (186) & 12.95 (109) & 20.89 (69) \\
2 & 5--9 & 18.46 (78) & 16.99 (44) & 18.48 (31) & 23.55 (38) \\
3 & 10--14 & 23.19 (26) & 21.89 (14) & 24.28 (8) & 28.43 (12) \\
4 & 15--19 & 26.46 (12) & 26.77 (5) & 25.36 (6) & 33.12 (4) \\
5 & 20--24 & 29.58 (6) & 27.73 (4) & 26.70 (5) & 36.74 (2) \\
6 & 25--29 & 32.61 (3) & 30.13 (2) & 29.74 (2) & 38.34 (1) \\
7 & 30--34 & 34.94 (2) & 31.35 (2) & 31.72 (1) & 38.93 (1) \\
8 & 35--39 & 36.41 (1) & 32.35 (1) & 33.10 (1) & 39.69 (1) \\
9 & 40--44 & 37.18 (1) & 33.36 (1) & 33.95 (1) & 40.36 (1) \\
10 & 45--49 & 37.37 (1) & 33.40 (1) & 33.32 (1) & 39.30 (1) \\
\midrule
\multicolumn{2}{@{}l}{7 windows of steps 15--49: max / min MSE} & 12.3$\times$ & 4.6$\times$ & 7.2$\times$ & 5.3$\times$ \\
\bottomrule
\end{tabular}}
\end{table*}

\subsection{One-step error with all layers sparse}
\label{app:C2}

At every step of the first and last windows, the one-step error with all layers sparse at 0.30 is measured and averaged over the window.
The first/last-window ratio is 6.4$\times$, 2.4$\times$, 31$\times$ and 0.18$\times$ on 1.3B, 14B, HunyuanVideo and Wan2.2 (on Wan2.2 the two windows belong to two different experts).
The corresponding final-video ratios are 463$\times$, 186$\times$, 69$\times$ and 109$\times$, so the one-step error understates them by 72$\times$, 78$\times$, 2.3$\times$ and about 600$\times$.
Measuring the amplification share as $1-\log(\text{one-step ratio})/\log(\text{final-video ratio})$ gives 0.70 on 1.3B and 0.83 on 14B: the gap between steps comes mainly from amplification in subsequent denoising.
On HunyuanVideo it is 0.19: the gap is mostly already in the one-step error.

\subsection{Local variation and attention concentration}
\label{app:C3}

PASA allocates steps by the L1 difference between the velocity predictions of adjacent steps (the local variation $\nu_s$). On the dense trajectory it is U-shaped, and its Spearman correlation with the final-video error of each window is $+0.02$.
ScalingAttention attributes the differences between steps to attention becoming increasingly concentrated along denoising. In the same runs as \S\ref{app:C2}, the first/last-window ratio of Hellinger$^2$ is 2.72$\times$ (0.112 / 0.041), and that of the density needed to cover 90\% of the attention mass is 2.20$\times$ (0.423 / 0.192).
Both are smaller than the 6.4$\times$ of the one-step error, and far smaller than the 463$\times$ of the final video.

\subsection{Comparison differing only in step-wise allocation}
\label{app:C4}

The two settings give all layers the same density in every step and have the same budget, differing only in step-wise allocation: one allocates density across steps using the trajectory gain; the other keeps the first 15 steps dense and uses a single density for all remaining steps.
At the three 1.3B 480p budgets (0.20 / 0.30 / 0.45), the former is higher by \dci{+1.77}{+1.46}{+2.08}, \dci{+3.18}{+2.62}{+3.74} and \dci{+4.46}{+3.73}{+5.18}\,dB, each 31/1, with wall-clock time differing by $\le$1.3\%.
On 14B 480p it is higher by \pd{+3.00}{+2.70}{+3.29}{32/0} (wall-clock time 264.5 / 269.9\,s).
LPIPS is better in all four groups (28/4--32/0).

\section{Observation 3: Across Inputs}
\label{app:D}

For each video, the per-layer cell probes ($\dens{=}0.30$) are measured once separately, and their layer ranking is compared with that of the mean over the 4 probing videos.
On 32 videos generated by 1.3B, the median Spearman correlation ($t$ = 0.85 / 0.45 / 0.15) is 0.84 / 0.91 / 0.90, close to the agreement among these videos themselves (0.93 / 0.95 / 0.95). At $t{=}0.85$ the most sensitive layer is layer~0 for 32/32 videos on 1.3B and 64/64 on 14B.
On 64 OpenVid videos per model, excluding the 4 probing videos among them gives 0.80 / 0.87 / 0.85 on 14B and 0.91 / 0.89 / 0.81 on HunyuanVideo, within 0.01 of the values before exclusion.

Across steps, the leave-one-out Spearman correlation between each prompt's ranking of the 10 windows and the ranking of the mean over the other prompts (on log MSE) has a median of 1.00 on 1.3B (0.99 with SpargeAttn selection), 0.98 on 14B, 0.99 on Wan2.2 and 0.95 on HunyuanVideo.
The first/last-window MSE ratio varies considerably across prompts (1.3B median 363$\times$), but the ranking is stable.

\begin{figure*}[t]
\centering
\includegraphics[width=\textwidth]{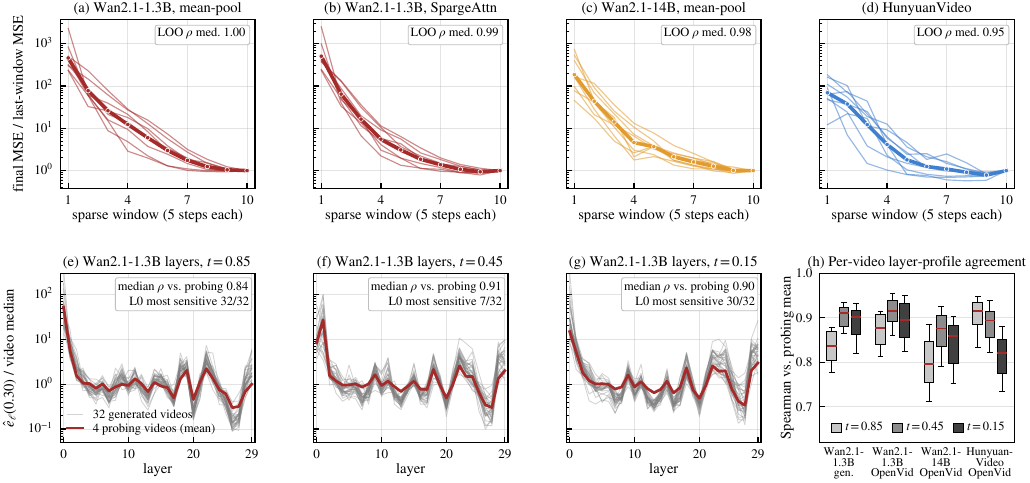}
\caption{Sensitivity is stable across inputs. (a--d)~Final-video MSE when only one window is sparsified (8 prompts each; boxes: median leave-one-out Spearman correlation). (e--h)~Per-video cell probes and layer-ranking Spearman on 1.3B and OpenVid videos.}
\label{fig:D1}
\end{figure*}

\section{Derivation and Verification of the Estimate}
\label{app:E}

The estimate relies on two approximations, separability and additivity (precise forms in \S\ref{app:E1}). Both are supported for rankings, but magnitudes remain approximate.
The ratio ``sparse together $\div$ sum of each alone'' in \S\ref{app:E4} tests the first-order superposition and the orthogonality parts at once.

\subsection{First-order error propagation}
\label{app:E1}

Let the sampler update of step~$s$ be $x_{s+1}=\Phi_s(x_s,v_s)$ ($s{=}0$ is the noisiest step). The sparse approximation adds an error $\epsilon_s$ to the velocity prediction $v_s$ at this step.
Linearizing around a reference dense trajectory, the linearized deviation satisfies $\delta_{s+1}=A_s\delta_s+B_s\epsilon_s$ ($A_s$ is the total Jacobian of one denoising step with respect to the state, including the chain term through $v_s$; $B_s$ is the step-size coefficient). Writing $\delta\equiv\delta_N$, unrolling from $\delta_0=0$ yields the linear term in Eq.~\eqref{eq:deviation}; for a one-step sampler, $P_s=A_{N-1}\cdots A_{s+1}B_s$.
The UniPC sampler~\cite{unipc} used for the Wan models is a multistep sampler whose update also depends on past velocities. Linearizing after including the past velocities in the state leaves the first-order representation $\delta\approx\sum_sP_s\epsilon_s$ unchanged; $P_s$ is no longer the product above.
At first order, VAE decoding replaces $P_s$ by $J_{\text{dec}}P_s$.
Below, expectations are over the probe inputs, and $P_s$ takes its value on the reference trajectory.
For a fixed model and velocity-tensor shape, the probe MSE $\eprobe_\lay(t_s,\dens)$ estimates $\mathbb E\lVert\epsilon_{\lay s}\rVert^2$ up to a common per-element normalization, which is absorbed into $w(s)$.

\paragraph{Precise form of the two approximations.}
Additivity has two parts: within one step, the perturbations of the layers superpose at first order, $\epsilon_s=\sum_\lay\epsilon_{\lay s}$; and the errors of different cells are orthogonal in expectation, $\mathbb E[\epsilon_{\lay s}\epsilon_{\lay's'}^\top]=0$ for $(\lay,s)\ne(\lay',s')$.
We impose the full cross-moment condition rather than scalar uncorrelatedness. It makes the propagated cross term $\mathbb E[\epsilon_{\lay s}^\top P_s^\top P_{s'}\epsilon_{\lay's'}]=\mathrm{tr}\big(P_s^\top P_{s'}\,\mathbb E[\epsilon_{\lay's'}\epsilon_{\lay s}^\top]\big)$ vanish for every pair of propagation operators and allows systematic bias in sparse-attention errors.
Separability holds if, within one step, the error second-moment matrices of the layers differ only by scale: $\Sigma_{\lay s}\equiv\mathbb E[\epsilon_{\lay s}\epsilon_{\lay s}^\top]=\alpha_{\lay s}\bar\Sigma_s$, $\mathrm{tr}\,\bar\Sigma_s=1$.
Then $\mathbb E\lVert P_s\epsilon_{\lay s}\rVert^2=\mathrm{tr}(P_s^\top P_s\Sigma_{\lay s})=\alpha_{\lay s}\,\mathrm{tr}(P_s^\top P_s\bar\Sigma_s)$, while $\mathbb E\lVert\epsilon_{\lay s}\rVert^2=\mathrm{tr}\,\Sigma_{\lay s}=\alpha_{\lay s}$, so their ratio $g(s)=\mathrm{tr}(P_s^\top P_s\bar\Sigma_s)$ does not depend on $\lay$. The isotropic case $\bar\Sigma_s=I/d$ ($d$ the latent dimension) is a special case, with $g(s)=\lVert P_s\rVert_F^2/d$.
Equation~\eqref{eq:additive} motivates ALLOT's factorized sensitivity estimate.
Proposition~1 establishes exact optimization of the empirically calibrated surrogate objective in Eq.~\eqref{eq:opt} at the attained budget.
A sufficient condition for separability is that the layers at a given step share the same normalized error second moment, while $P_s$ can remain anisotropic. This distinction matters because low-frequency velocity perturbations have a larger effect on the final video (Table~1 of HEART~\cite{heart}).

The gain $g(s)$ therefore depends on both the trajectory and the shared error structure. ALLOT captures this step dependence with a trajectory gain calibrated directly from sparse-window generations, setting $w(s)=M_k/E_k$ for $s\in\Omega_k$. The calibration also captures average cross-layer interactions and latent-to-pixel conversion. The oracle diagnostic in \S\ref{app:E7} shows why the actual sparse error is used: energy-preserving or energy-matched transformations produce different final-video errors. The resulting allocation is evaluated in \S\ref{app:E5} and \S\ref{sec:exp:ablation}.

\subsection{Proof of Eq.~\texorpdfstring{\eqref{eq:additive}}{(additive estimate)} and Proposition~1}
\label{app:E2}

\paragraph{Eq.~\eqref{eq:additive}.}
Under additivity and separability, the squared energy of the linearized deviation takes the form in Eq.~\eqref{eq:additive}, with $g(s)$ depending only on the step~$s$.
Let $\delta^{(1)}=\sum_{\lay,s}P_s\epsilon_{\lay s}$ denote the linearized deviation in Eq.~\eqref{eq:deviation}. Then
\[
\mathbb E\lVert\delta^{(1)}\rVert^2=\sum_{(\lay,s),(\lay',s')}\mathrm{tr}\big(P_s^\top P_{s'}\,\mathbb E[\epsilon_{\lay's'}\epsilon_{\lay s}^\top]\big).
\]
By orthogonality the terms with $(\lay,s)\ne(\lay',s')$ vanish, leaving $\sum_{\lay,s}\mathrm{tr}(P_s^\top P_s\Sigma_{\lay s})$; by separability, $\Sigma_{\lay s}=\alpha_{\lay s}\bar\Sigma_s$, so each term equals $\alpha_{\lay s}\,\mathrm{tr}(P_s^\top P_s\bar\Sigma_s)=g(s)\,\mathbb E\lVert\epsilon_{\lay s}\rVert^2$, using $\mathbb E\lVert\epsilon_{\lay s}\rVert^2=\mathrm{tr}\,\Sigma_{\lay s}=\alpha_{\lay s}$.
Substituting gives Eq.~\eqref{eq:additive} at first order.\hfill$\blacksquare$

\paragraph{Proposition~1.}
Let $\dens^\lambda$ minimize $\hat S(\lay,s;\dens)+\lambda\lut(\dens)$ per cell over the candidate set $\mathcal B$, with $\lambda\ge0$.
Write $S_{\text{tot}}(\dens)=\sum_{\lay,s}\hat S(\lay,s;\dens_{\lay s})$ and $\lut_{\text{tot}}(\dens)=\sum_{\lay,s}\lut(\dens_{\lay s})$.
For any table $\dens$ with values in $\mathcal B$ and $\lut_{\text{tot}}(\dens)\le\lut_{\text{tot}}(\dens^\lambda)$, per-cell optimality gives $S_{\text{tot}}(\dens)+\lambda\lut_{\text{tot}}(\dens)\ge S_{\text{tot}}(\dens^\lambda)+\lambda\lut_{\text{tot}}(\dens^\lambda)$; rearranging,
\[
S_{\text{tot}}(\dens)\ \ge\ S_{\text{tot}}(\dens^\lambda)+\lambda\big[\lut_{\text{tot}}(\dens^\lambda)-\lut_{\text{tot}}(\dens)\big]\ \ge\ S_{\text{tot}}(\dens^\lambda).
\]
Thus $\dens^\lambda$ is the exact discrete optimum of Eq.~\eqref{eq:opt} over $\mathcal B$ at the attained budget $\lut_{\text{tot}}(\dens^\lambda)$~\cite{everett}.\hfill$\blacksquare$

\paragraph{Gap for a given budget $C$.}
With discrete candidate densities, not every budget is exactly attainable.
When the bisection terminates, it has a table $\dens^{\lambda_{\mathrm{hi}}}$ whose cost does not exceed $C$ and a table $\dens^{\lambda_{\mathrm{lo}}}$ whose cost exceeds $C$ (\S\ref{app:F5}).
Any table $\dens$ with cost $\le C$ also satisfies $\lut_{\text{tot}}(\dens)<\lut_{\text{tot}}(\dens^{\lambda_{\mathrm{lo}}})$, so by Proposition~1 $S_{\text{tot}}(\dens)\ge S_{\text{tot}}(\dens^{\lambda_{\mathrm{lo}}})$; the optimum under budget $C$ therefore lies between $S_{\text{tot}}(\dens^{\lambda_{\mathrm{lo}}})$ and $S_{\text{tot}}(\dens^{\lambda_{\mathrm{hi}}})$. Algorithm~1 returns one of the two: $\dens^{\lambda_{\mathrm{hi}}}$ is within their difference of this optimum, and $\dens^{\lambda_{\mathrm{lo}}}$ is no worse than it but exceeds $C$; the two tables usually differ in only one cell.
On 15 tables used in the comparisons of the main text and Appendices~\ref{app:G}--\ref{app:H}, this difference is 0.002\%--0.5\% of the estimated error.

\subsection{Final video with a single layer sparse over all steps}
\label{app:E3}

Each run sparsifies one of six representative layers over all 50 steps while keeping the other layers dense.
This tests the layer ranking predicted by
$\sum_s w(s)\hat e_\ell(t_s,\dens)$.
The measured ranking matches the prediction exactly on 1.3B and HunyuanVideo (Spearman 1.00) and nearly matches it on 14B (0.94), where only layers 19 and 39 swap.

On HunyuanVideo, coverage density instead ranks layers 54 and 58 as the two most sensitive, although sparsifying them yields 11.83 and 15.93\,dB higher PSNR than sparsifying layer~1.

\subsection{Separability and additivity}
\label{app:E4}

Table~\ref{tab:E1} summarizes the tests. The ratio ``sparse together $\div$ sum of each alone'' tests the first-order superposition and the orthogonality parts at once.

\begin{table*}[t]
\centering\ttiny
\caption{Tests of the two approximations.}
\label{tab:E1}
\begin{tabular}{@{}P{3.8cm}P{9.6cm}P{3.0cm}@{}}
\toprule
Test & Reading & Conclusion \\
\midrule
Separability: single layer sparse over all steps (\S\ref{app:E3}) & ranking of the 6 layers identical on 1.3B and HunyuanVideo, Spearman 0.94 on 14B; ratio between the most and the least sensitive layer predicted 9.3$\times$, measured 4.8$\times$ (1.3B), predicted 146$\times$, measured 7.7$\times$ (14B), predicted 286$\times$, measured 30$\times$ (HunyuanVideo) & ranking holds, magnitude compressed \\
Layer groups within one window: final MSE per unit of measured injected energy & at density 0.05, largest/smallest group ratio 2.17 [1.89, 2.50] in window~4 and 1.25 [0.91, 1.71] in window~6; energy-matched layer~0/comparison-group ratios in windows 1--2 are 0.56--0.82 ($n{=}8$) & point estimates within $2\times$ in window~6 and the energy-matched early windows; 2.17$\times$ in window~4 \\
Across steps: two windows sparse together $\div$ sum of each alone & over 7 window pairs, 1.3B 0.86--1.26 and 14B 0.83--1.36; adjacent windows within steps 15--34 super-additive by 1.16--1.36$\times$ (4 of 6 pairs at $p\le0.04$, $n{=}8$) & approximately additive \\
Across layers: all layers sparse together $\div$ sum of each layer alone & 1.3B 6.5 at high noise to 1.1 at low noise, 14B 1.9 to 0.8, HunyuanVideo 0.74--0.89 to 1.2--1.5; for two layers sparse together within one step, median ratio $1\pm0.08$ & not additive in magnitude; absorbed by $w$ \\
Additive prediction of isolated probes vs.\ marginals measured in context & Spearman: 1.3B 0.63 in the deployed table's context and 0.36 in a uniform 0.15 context; HunyuanVideo 0.77 in the deployed table's context & less accurate in sparser contexts \\
Additive prediction vs.\ final video for arbitrary tables (80 tables, \S\ref{app:E5}) & Kendall $\tau$: pooled 0.83 ($R^2$ 0.90), within one budget 0.78 (per budget 0.70 / 0.77 / 0.86); by family of tables: convex combinations 0.91, jitters 0.82, random tables 0.72, log--log slopes 0.98--1.14 / 0.48--0.75 / 0.26--0.44 & rankings within one budget usable; weaker for tables further from the solved ones \\
\bottomrule
\end{tabular}
\end{table*}

\paragraph{Direct layer-group check.}
On 1.3B, we sparsify one of three disjoint layer groups within a five-step window and divide each video's final MSE by the group's measured injected energy summed over that window ($n{=}8$; paired log-ratio intervals). The injected energy is the mean MSE of the two CFG branches before guidance, summed over the perturbed steps.
At density 0.05, the groups are layer~0, layers~1 and~19 (plus layer~22 in window~6), and all remaining layers; layer~0 in window~6 produces too little final error for a stable ratio and is excluded. In windows 1--2, sparsifying layer~0 at density 0.30 drives the final-video error toward saturation, so we raise its density to 0.40--0.50 to bring its injected energy within $1.8\times$ of the comparison groups.
In point estimates, the measured ratios of final MSE to injected energy differ by less than $2\times$ in window~6 and the energy-matched early-window comparisons; window~4 shows a 2.17$\times$ group spread. The test covers only these groups and densities.

\subsection{Prediction vs.\ measurement for arbitrary tables}
\label{app:E5}

On 1.3B 480p$\times$33 we construct 80 tables (log-convex combinations of ten existing tables at the deployed budget, random tables with a log-uniform density per block of 5 layers $\times$ 5 steps, and log-normal jitters of the solved table), run each on 4 held-out prompts, and compare the additive-model prediction with the final-video pixel MSE.

\begin{figure}[!htbp]
\centering
\includegraphics[width=\columnwidth]{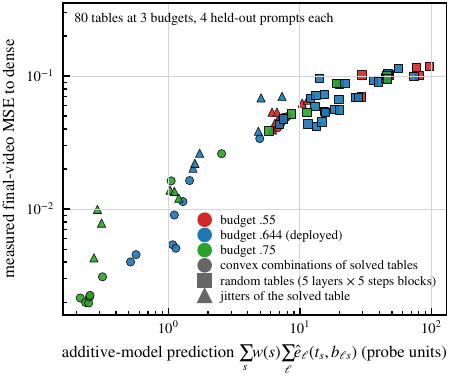}
\caption{Additive-model prediction vs.\ final-video MSE for 80 tables at three budgets (color: budget; marker: family of the table; 4 held-out prompts each). Kendall $\tau$ per budget and pooled in the last row of Table~\ref{tab:E1}.}
\label{fig:E1}
\end{figure}

\subsection{Reassigning the same densities}
\label{app:E6}

The setting is the same as in \S\ref{app:I4} (1.3B 480p$\times$33, mean-pooled top-$k$, VSA, $n{=}32$).
The densities of ALLOT's table are reassigned unchanged: the multiset is the same and the cost identical. Only which cell gets which density changes.
We test four types of reassignment, yielding five reassigned tables: shuffling layers within each step, shuffling steps within each layer, fully random reassignment with two seeds, and assignment using the reversed estimate (cells with higher estimated sensitivity receive lower density).

\begin{table*}[t]
\centering\footnotesize
\renewcommand{\arraystretch}{1.15}
\caption{Reassigning the same densities ($\Delta$ = row $-$ ALLOT).}
\label{tab:E2}
\begin{tabular*}{\textwidth}{@{\extracolsep{\fill}}lccc@{}}
\toprule
Assignment & PSNR & $\Delta$PSNR & Additive-model prediction \\
\midrule
ALLOT & 25.99 & -- & 0.37 \\
Shuffle layers within each step & 22.60 & \pd{-3.40}{-3.94}{-2.86}{0/32} & 1.07 \\
Shuffle steps within each layer & 13.77 & \pd{-12.23}{-13.60}{-10.85}{0/32} & 10.6 \\
Fully random (seed 1) & 10.21 & \pd{-15.78}{-17.48}{-14.09}{0/32} & 45.6 \\
Fully random (seed 0) & 9.82 & \pd{-16.17}{-17.85}{-14.49}{0/32} & 56.1 \\
Reversed estimate & 7.95 & \pd{-18.05}{-19.65}{-16.44}{0/32} & 146 \\
\bottomrule
\end{tabular*}
\end{table*}

All five reassignments are 0/32 on all three metrics, and the measured order matches the order predicted by the additive model (Kendall $\tau=1.0$).
Shuffling layers within each step is still 3.40\,dB lower, of the same order as the drop from ``same density for all layers'' ($-3.22$\,dB), showing that the layer ranking has an effect of its own.

\subsection{Oracle diagnostic of trajectory calibration}
\label{app:E7}

\paragraph{Protocol.}
To test whether a generic perturbation reproduces the downstream effect of the actual sparse-attention error, we modify that error within one five-step window. At each run's current latent state $x_s$, the sampler receives
\begin{equation}
\small
\begin{aligned}
  v_s^{\mathrm{dense}}(x_s)+T_s\!\left(\epsilon_s(x_s)\right), \quad \epsilon_s(x_s)=v_s^{\mathrm{sparse}}(x_s)-v_s^{\mathrm{dense}}(x_s).
  \end{aligned}
\end{equation}
Identity $T_s(\epsilon)=\epsilon$ reproduces the ordinary sparse-window run. A random-sign transform independently flips each element of $\epsilon_s$, preserving its magnitude, while a Gaussian control replaces $\epsilon_s$ with independent noise matched to its mean-square magnitude after classifier-free guidance. All other steps remain dense.
These oracle interventions require the dense velocity and need not correspond to realizable attention masks; ALLOT itself uses only ordinary sparse-window generations. Because each transform is applied at its run's current state, the experiment measures end-to-end responses to changes in sparse-error structure rather than isolated linear propagation gains.
The top-$k$ and fixed sliding-window masks have densities 0.30 and 0.304, respectively.

\begin{table}[!htbp]
\centering\ttiny
\caption{Oracle diagnostic: mean final-video PSNR (dB) to each model's dense reference. All rows use 8 prompts, 480p$\times$33, 50 steps and seed~0; windows~1 and~5 cover steps 0--4 and 20--24. Only the indicated window is perturbed.}
\label{tab:E3}
\suppfit{\begin{tabular}{@{}llccc@{}}
\toprule
Model and mask & Window & Sparse & Random signs & Gaussian \\
\midrule
1.3B, top-$k$ & 1 & 10.71 & 29.57 & 25.68 \\
1.3B, top-$k$ & 5 & 29.58 & 38.55 & 38.64 \\
1.3B, sliding window & 1 & 9.16 & 28.23 & 26.66 \\
1.3B, sliding window & 5 & 26.21 & 38.23 & 38.48 \\
14B, top-$k$ & 1 & 10.70 & 25.93 & 26.54 \\
14B, top-$k$ & 5 & 27.73 & 37.17 & 37.45 \\
\bottomrule
\end{tabular}}
\end{table}

\paragraph{Result.}
Across all six settings, both oracle controls yield higher PSNR to the dense reference than the ordinary sparse run for every one of the eight prompts, spanning both model sizes and both mask types (Table~\ref{tab:E3}). This consistent direction shows that the downstream response depends on the perturbation pattern, supporting the use of actual sparse-window errors when measuring $M_k$. The layer-group and independent-table tests in \S\ref{app:E4}--\ref{app:E5} evaluate weight sharing and allocation ranking separately.

\paragraph{Scope.}
The diagnostic spans six Wan2.1 settings across two model sizes, two mask types and two windows, with eight prompts per setting. Corresponding interventions on Wan2.2 and HunyuanVideo remain untested. The first window is nonlinear and approaches saturation, so these results characterize pattern sensitivity rather than a universal propagation gain.

\section{Method Details: Cell Probes, In-Context Check, Latency Tables, Solving and Integration}
\label{app:F}

\subsection{Cell probes: definition, grid and interpolation}
\label{app:F2}

Cell probe $\eprobe_\lay(t,\dens)$: only layer~$\lay$ is sparsified to density~$\dens$ with the selection strategy used at deployment, the other layers stay dense, one teacher-forced forward pass is run at noise level~$t$, and the MSE between the velocity prediction and that of the dense model is taken.
The inputs are the VAE latents of 4 OpenVid real videos with noise added at level~$t$.
Different selection strategies have a layer-ranking correlation of only 0.48--0.83 at the same nominal density, so probes are by default measured under the selection strategy used at deployment.

The mean-pooled top-$k$ probes are measured at 8 densities (0.05--0.60) $\times$ 9 noise levels (0.85--0.05).
They are interpolated onto the candidate densities and steps as in line~2 of Algorithm~1 (\S\ref{app:F5}).

\subsection{In-context check}
\label{app:F3}

The cell probes are measured with the other layers dense, whereas at deployment the other layers are also sparse. The in-context check examines this difference.
With all layers at a uniform $\dens_0{=}0.30$ as the context, a group of layers (layer~0, layer~1, layers 0+1, or one depth band) is restored to dense, and the actual reduction of this step's velocity error is divided by the sum of this group's cell probes (8 OpenVid videos including the 4 probing videos, one value per noise level):
\[
\kappa=\frac{e(\text{all layers at }\dens_0)-e(\text{this group dense, others at }\dens_0)}{\sum_{\lay\in\text{group}}\eprobe_\lay(t,\dens_0)},
\]
where $e(\cdot)$ is this step's one-step error under the setting in parentheses.
$\kappa\approx1$ means the isolated measurement still holds in context; $\kappa\ll1$ means the probes overestimate the role of this group in context, and $\kappa\gg1$ means that they underestimate it.

$\kappa$ is far from 1, in model-specific ways: layer~0 is at 1.9--15.5 on Wan2.1-1.3B for $t\ge0.35$ and at 0.09--0.44 in the high-noise expert of Wan2.2, and on HunyuanVideo restoring layers 0+1 removes at most 25\% of the in-context error.
We tested the correction below on final videos. It raises PSNR on HunyuanVideo (\pd{+0.46}{+0.23}{+0.69}{100/60}, $n{=}160$; LPIPS and DISTS not significant) but lowers it by 6.09\,dB on Wan2.1-1.3B and by 0.78\,dB on both Wan2.2 experts (0/32 each). Since its effect does not carry over across models, no table in the paper is corrected.
Against the uniform baseline at equal density ($n{=}160$), the uncorrected HunyuanVideo table gives a non-significant \pd{+0.27}{-0.04}{+0.57}{95/65} and the corrected one \pd{+0.73}{+0.47}{+0.99}{111/49}.

\paragraph{Correction.}
Each layer takes the $\kappa$ of its group, $c_\lay(t)$: layers~0 and~1 their own, and every other layer that of its depth band, the model's layers being cut into six bands of consecutive layers. $c_\lay$ is clipped to $[0.25,8]$ (a negative $\kappa$, where restoring the group increases the error, counts as 0.25) and rescaled within each noise level so that $\sum_\lay c_\lay\eprobe_\lay(t,\dens_0)=\sum_\lay\eprobe_\lay(t,\dens_0)$. It is then interpolated log-linearly in $t$ onto each step and multiplies the cell probes at every candidate density, $\eprobe'_\lay(t_s,\dens)=c_\lay(t_s)\,\eprobe_\lay(t_s,\dens)$. It thus changes only the layers' shares within one step: each step's total remains calibrated by the trajectory gain, and the latency table is unchanged.

\subsection{Latency tables of each path}
\label{app:F4}

The latency table $\lut(\dens)$ is the time of one attention call at density~$\dens$ divided by that of dense FlashAttention on the same shape.
In the solver, $\lut$ is clipped to 1.

\begin{figure*}[t]
\centering
\includegraphics[width=\textwidth]{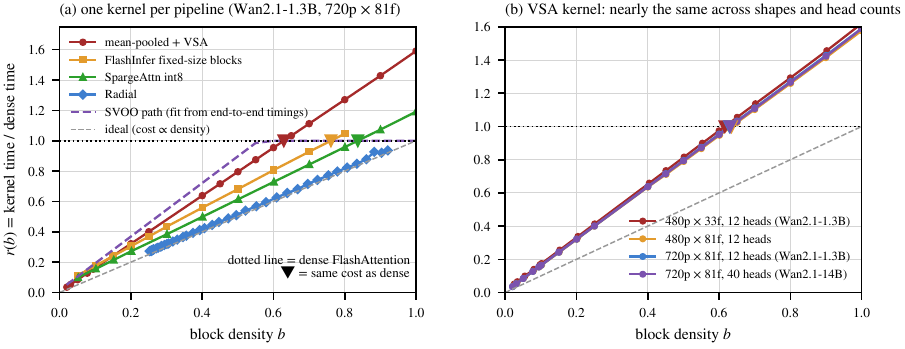}
\caption{Latency tables. (a)~Kernels of each pipeline on Wan2.1-1.3B 720p$\times$81; $\blacktriangledown$ marks the density at which cost equals dense. (b)~The VSA kernel is nearly identical across shapes and head counts.}
\label{fig:F2}
\end{figure*}

Only changing the kernel moves the crossover. Cells above the crossover are set to dense, and assuming cost proportional to density or FLOPs would overestimate the benefit of sparsity.

\subsection{Solving: Algorithm~1, candidate densities and density cap}
\label{app:F5}

\par\medskip\noindent
\fbox{\begin{minipage}{\dimexpr\columnwidth-2\fboxsep-2\fboxrule\relax}
\small\textbf{Algorithm 1. Latency-constrained density allocation}\par\smallskip
\ttiny
\textbf{Input:} cell probes $\eprobe_\lay(t,\dens)$ (measured noise levels $T$, densities $D$), trajectory gain $w(s)$, latency table $\lut(\dens)$, budget $C$, density cap $\dens_{\max}$.
\begin{enumerate}[leftmargin=1.6em,label=\arabic*:]
\item Candidate densities $\mathcal{B}$: 200 uniform points on $[0.05,\min(\dens_{\max},0.99)]$, $\cup\,\{1\}$.
\item For each cell $(\lay,s)$: enforce $\eprobe$ non-decreasing from high to low density; interpolate linearly in $\log\dens$--$\log\eprobe$ onto $\mathcal{B}$, and linearly in $t$ onto $t_s$ (endpoint values when $t_s$ is outside $T$).
\item $\hat S(\lay,s;b)=w(s)\cdot\eprobe_\lay(t_s,b)$, with $0$ for dense ($b{=}1$); $R(b)=\min(\lut(b),1)$, with $1$ for dense.
\item Take $\lambda_{\mathrm{lo}}$, $\lambda_{\mathrm{hi}}$ (initially $[10^{-12},10^{3}]$) such that the total cost is above and below $C$, respectively.
\item Repeat 80 times: $\lambda=\sqrt{\lambda_{\mathrm{lo}}\,\lambda_{\mathrm{hi}}}$;
\item \quad for every cell independently, $b^*(\lay,s)=\argmin_{b\in\mathcal{B}}\,\hat S(\lay,s;b)+\lambda\,R(b)$;
\item \quad if $\sum_{\lay,s}R(b^*(\lay,s))>C$ then $\lambda_{\mathrm{lo}}\leftarrow\lambda$, else $\lambda_{\mathrm{hi}}\leftarrow\lambda$.
\item Return the table $b^*$ at $\lambda=\sqrt{\lambda_{\mathrm{lo}}\,\lambda_{\mathrm{hi}}}$ (its cost may exceed $C$ by one marginal cell).
\end{enumerate}
\end{minipage}}
\par\medskip
$\dens_{\max}$ is the highest probed density of the selection strategy (0.57--0.90); on HunyuanVideo at equal density (\S\ref{app:J1}) it is the uniform density, 0.30.

\subsection{Integration}
\label{app:F6}

ALLOT keeps the selection strategy's scoring and ranking, and the kernel is the same as the baseline's (XAttention's official kernel is replaced by FlashInfer on both sides, see \S\ref{app:G2}; for Radial's kernel paths see \S\ref{app:H2}); only each cell's cutoff is set to the density in the table.
\begin{itemize}[nosep,leftmargin=1.5em]
\item \textbf{XAttention:} keeps blocks by block score up to the table's density, replacing its per-head cumulative threshold.
\item \textbf{SVOO:} keeps key clusters in order of centroid score until the fraction of key tokens reaches the table value, bypassing its top-$p$ threshold, its per-(layer, head) density clamping and its dense warm-up steps and first layer.
\item \textbf{SVG2:} likewise keeps key clusters in its ranking order until the fraction of key tokens reaches the table value, bypassing its top-$p$ threshold, its minimum key-cluster ratio and its first-steps dense rule.
\item \textbf{SpargeAttn:} uses its built-in top-$k$ entry.
\item \textbf{Radial:} takes the nearest reachable \texttt{decay\_factor} that does not exceed the table value.
\end{itemize}
Cells marked dense in the table all use the exact dense path.

\paragraph{Effect of density truncation.}
The effect of replacing threshold truncation with density truncation depends on the selection strategy.
On the SVOO 480p$\times$33 pipeline, we preserve its default per-cell densities and dense-cell rules (\S\ref{app:B2}) and change only the truncation rule.
This lowers PSNR by 3.00\,dB on 1.3B (\pd{-3.00}{-3.53}{-2.46}{0/32}) and by 1.96\,dB on HunyuanVideo (\pd{-1.96}{-2.69}{-1.24}{4/28}).

For XAttention, density truncation has the opposite effect.
At equal wall-clock time, ALLOT with density truncation exceeds its threshold-based counterpart by 4.50\,dB (\pd{+4.50}{+4.00}{+5.01}{32/0}).
The threshold-based variant assigns each cell a cumulative-mass threshold with the same cost as truncating that cell at its mean density.
For a head with concentrated attention, a low threshold may retain only one or two blocks.
Replacing each threshold in XAttention's default allocation with the density it induces raises PSNR by 2.27\,dB (\pd{+2.27}{+1.69}{+2.85}{30/2}) at 1.9\% more wall-clock time.
Even against this strengthened default, ALLOT remains 5.98\,dB higher at equal wall-clock time (\pd{+5.98}{+5.46}{+6.50}{32/0}; \S\ref{app:G2}).

\subsection{Offline cost}
\label{app:F7}

\begin{table}[!htbp]
\centering\ttiny
\caption{Offline cost per model (GPU-h, including model loading).}
\label{tab:F2}
\suppfit{\begin{tabular}{@{}lcccc@{}}
\toprule
Model & Cell probes & Trajectory gain & Latency table & Total \\
\midrule
Wan2.1-1.3B & 2.8 & 3.1 & minutes & $\approx$6 \\
Wan2.1-14B & 17.4 & shared with 1.3B & minutes & $\approx$17 \\
HunyuanVideo & 18.9 & 3.8 & minutes & $\approx$23 \\
\bottomrule
\end{tabular}}
\end{table}

On Wan2.2-A14B the cell probes of the two experts take 26\,GPU-h and its own trajectory gain 6.5\,GPU-h of generation.
When switching the selection strategy, we re-measure the cell probes (3.0\,GPU-h each for SVOO and SpargeAttn on 1.3B, 22.6\,GPU-h for Radial on the 720p latent, and 57\,GPU-h for Radial on HunyuanVideo) and the latency table (minutes), and reuse the trajectory gain measured with mean-pooled top-$k$ (Appendix~\ref{app:K}).

\section{Comparison with Other Allocation Methods}
\label{app:G}

\subsection{Paired intervals}
\label{app:G1}

Table~\ref{tab:G1} gives the paired differences in all four metrics against the default and HEART at 480p$\times$81 (Table~\ref{tab:compare}) and 720p$\times$81 (\S\ref{app:G5}), with verdicts as in Appendix~\ref{app:A}; the other baselines are compared in Table~\ref{tab:G2}.

\begin{table*}[t]
\centering\readablepairedtable
\caption{Paired differences (ALLOT $-$ baseline) for the XAttention rows default, EBC and HEART (EBC + TMR) of Table~\ref{tab:compare} and XAttention on Wan2.1-1.3B 720p$\times$81 (\S\ref{app:G5}); verdicts follow \S\ref{app:A1}.}
\label{tab:G1}
\begin{tabular*}{\textwidth}{@{\extracolsep{\fill}}P{2.8cm}P{2.2cm}cccccc@{}}
\toprule
Setting & Baseline & $n$ & $\Delta$PSNR & $\Delta$SSIM & $\Delta$LPIPS & $\Delta$DISTS & Verdict \\
\midrule
XAttention 1.3B 480p$\times$81 (ALLOT at the default's wall-clock time) & default & 32 & \pd{+7.23}{+6.60}{+7.85}{32/0} & \pd{+0.176}{+0.148}{+0.204}{32/0} & \pd{-0.163}{-0.187}{-0.138}{32/0} & \pd{-0.059}{-0.073}{-0.044}{31/1} & win \\
XAttention 1.3B 480p$\times$81 & HEART (EBC) & 32 & \pd{+6.96}{+6.35}{+7.56}{32/0} & \pd{+0.155}{+0.127}{+0.183}{32/0} & \pd{-0.132}{-0.154}{-0.111}{32/0} & \pd{-0.045}{-0.055}{-0.035}{31/1} & win \\
XAttention 1.3B 480p$\times$81 (TMR on both sides) & HEART (EBC + TMR) & 32 & \pd{+6.59}{+5.99}{+7.19}{32/0} & \pd{+0.142}{+0.115}{+0.168}{32/0} & \pd{-0.120}{-0.141}{-0.099}{32/0} & \pd{-0.040}{-0.050}{-0.030}{31/1} & win \\
XAttention 1.3B 720p$\times$81 (ALLOT at the default's wall-clock time) & default & 32 & \pd{+6.35}{+5.61}{+7.09}{32/0} & \pd{+0.110}{+0.092}{+0.127}{32/0} & \pd{-0.141}{-0.165}{-0.117}{32/0} & \pd{-0.062}{-0.078}{-0.046}{29/3} & win \\
XAttention 1.3B 720p$\times$81 (ALLOT at the default's wall-clock time) & HEART (EBC) & 32 & \pd{+5.16}{+4.40}{+5.92}{32/0} & \pd{+0.079}{+0.065}{+0.093}{32/0} & \pd{-0.095}{-0.114}{-0.076}{31/1} & \pd{-0.032}{-0.045}{-0.019}{27/5} & win \\
XAttention 1.3B 720p$\times$81 (TMR on both sides) & HEART (EBC + TMR) & 32 & \pd{+5.21}{+4.44}{+5.98}{32/0} & \pd{+0.083}{+0.070}{+0.097}{32/0} & \pd{-0.100}{-0.120}{-0.079}{31/1} & \pd{-0.035}{-0.048}{-0.022}{28/4} & win \\
\bottomrule
\end{tabular*}
\end{table*}

\subsection{XAttention and HEART}
\label{app:G2}

\paragraph{Setting.}
Following the setting of HEART's ablation~\cite{heart} (Wan2.1-1.3B, 832$\times$480$\times$81, 50 steps, classifier-free guidance (CFG~\cite{cfg}) scale 5, UniPC, $n{=}32$), all settings keep the first 5 steps dense.
Selection uses the official block estimation~\cite{xattention}. Attention uses the FlashInfer fixed 128-block block-sparse kernel~\cite{flashinfer}. On real inputs the two produce identical block masks.
The budget is the cost of the default setting under the same latency table, 0.734 of dense.

\paragraph{HEART's EBC.}
We re-implement EBC from HEART's description, without TMR (cross-step mask reuse): only one head at a time is sparsified to one of the thresholds $\{0.85,0.90,0.95\}$, one step is sampled from each of four step segments to measure the frequency-weighted error of the CFG-combined velocity, and an integer program then picks one threshold per head such that the mean actual sparsity is no lower than that of a shared threshold of 0.90. Calibration uses 10 prompts that do not overlap with the evaluation set.
EBC gives \pd{+1.29}{+1.00}{+1.57}{32/0} over the default.

\paragraph{HEART (EBC + TMR).}
Following HEART's \S4.2 and Algorithm~7.3, we implement TMR, which reuses masks across steps: for each layer, CFG branch and head, the token-wise means of query and key at the last re-scoring are stored together with that step's block scores and mask; if the sum of the L1 drifts of the two at the current step does not exceed $\delta_{\mathrm{TMR}}{=}30$ (HEART's default), the previous result is reused, otherwise the head is re-scored.
When fewer than 40\% of a layer's heads need re-scoring, the whole layer reuses its masks; when more than 80\% need it, the whole layer is re-scored; otherwise, heads are handled individually (HEART's values).

\paragraph{ScalingAttention (FAST).}
On EBC's 10 calibration trajectories, FAST measures, for each layer, head and candidate density, the attention mass within the selected blocks and converts it to a Hellinger distance. A three-parameter fidelity surface $A=\beta[1-(1-x)^\gamma]$ then sets a fidelity requirement for each (step, layer), where $x=(\ell'+\omega t')/(1+\omega)$ and $\ell',t'\in[0,1]$ are the normalized layer depth and noise level (1 for shallow layers and early steps), so the requirement is higher for shallow layers and early steps, following the description in its main text; each head takes the sparsest density meeting it.
$\gamma$ and $\omega$ are grid-searched, $\beta$ is set by bisection to meet the budget, and the combination maximizing the minimum fidelity is selected ($\gamma{=}4$, $\omega{=}1$).
$\gamma{=}4$ lies on the edge of the grid. Extending $\gamma$ to 64, the table selected by the same criterion is 1.94\,dB lower in PSNR, so $\gamma{=}4$ is kept.
Deviations from the original: its static topology and kernel are not reproduced, and a 5-step warm-up is added.
ALLOT is \pd{+3.63}{+3.04}{+4.23}{31/1}.

\paragraph{PASA.}
PASA keeps the first 10 steps dense, then gives each step one density shared by all layers, proportional to the local variation $\nu_s$, with the coefficient set by bisection to meet the budget.
ALLOT is \pd{+4.47}{+4.14}{+4.81}{32/0}.

\paragraph{CalibAtt-style and DFSAttn-style rules.}
The exponential threshold curve takes $\theta_\infty{=}0.842$ and $\theta_0{=}0.99$ from the appendix of CalibAtt~\cite{calibatt}; the three levels of DFSAttn~\cite{dfsattn} are multiplied by one common factor of 1.52 to match the budget.

\paragraph{DiTFastAttn-style rule.}
Following Algorithm~1 of DiTFastAttn~\cite{ditfastattn}, each step after the warm-up visits the layers in order and gives layer~$\ell$ the sparsest density for which the sum of the cell probes chosen so far in that step stays below $\delta_{\mathrm{DFA}}(\ell{+}1)/L$, and dense otherwise. The probes and latency table are ALLOT's, without the trajectory gain, and $\delta_{\mathrm{DFA}}$ is set by bisection to meet the budget.

\begin{table*}[t]
\centering\footnotesize
\setlength{\tabcolsep}{2pt}
\renewcommand{\arraystretch}{1.12}
\caption{XAttention allocation settings ($n{=}32$; shared latency table and budget). Top: construction and absolute metrics. Bottom: paired differences, $\Delta=\text{ALLOT}-\text{row}$. Default and CalibAtt-style use ALLOT at the default's time; HEART (EBC + TMR) uses ALLOT + TMR.}
\label{tab:G2}
\begin{tabular*}{\textwidth}{@{\extracolsep{\fill}}P{3.0cm}P{4.5cm}cccccc@{}}
\toprule
Setting & Construction & \hd{Dense\\cells} & \hd{Time\\(s)} & PSNR & SSIM & LPIPS & DISTS \\
\midrule
Dense & -- & 1500 & 183.8 & -- & -- & -- & -- \\
Default & 5 dense steps; threshold 0.90 & 150 & 148.7 & 18.14 & 0.659 & 0.268 & 0.154 \\
Default, density cutoff & Default per-cell densities; density cutoff & 150 & 151.5 & 20.41 & 0.745 & 0.181 & 0.111 \\
HEART (EBC) & Per-head thresholds; integer programming & 150 & 151.7 & 19.43 & 0.699 & 0.221 & 0.130 \\
HEART (EBC + TMR) & EBC + cross-step mask reuse & 150 & 145.0 & 19.54 & 0.706 & 0.214 & 0.128 \\
Efficient-vDiT & Per-layer density; layers 0--2 dense & 285 & 151.1 & 19.08 & 0.702 & 0.225 & 0.133 \\
DiTFastAttn-style~\cite{ditfastattn} & Greedy per step; error cap $\delta_{\mathrm{DFA}}(\ell{+}1)/L$, else dense & 208 & 152.2 & 19.38 & 0.708 & 0.219 & 0.130 \\
ScalingAttention (FAST) & Hellinger profile; three-parameter surface; per-head density & 159 & 152.9 & 22.75 & 0.790 & 0.141 & 0.117 \\
PASA & 10 dense steps; per-step density $\propto\nu_s$ & 360 & 151.8 & 21.91 & 0.768 & 0.163 & 0.117 \\
CalibAtt-style & Exponential threshold (\S\ref{app:G2}); $\beta{=}2.24$ & 150 & 149.6 & 19.67 & 0.700 & 0.222 & 0.140 \\
DFSAttn-style~\cite{dfsattn} & 12 dense steps; decrease every 12 steps & 360 & 151.3 & 24.86 & 0.834 & 0.107 & 0.096 \\
ALLOT & Solved & 587 & 151.3 & 26.39 & 0.854 & 0.089 & 0.085 \\
ALLOT + TMR & ALLOT's table + mask reuse & 587 & 146.0 & 26.13 & 0.847 & 0.094 & 0.089 \\
ALLOT (default time) & Solved at default time & 521 & 148.6 & 25.37 & 0.835 & 0.106 & 0.096 \\
\bottomrule
\end{tabular*}
\par\medskip
\begin{tabular*}{\textwidth}{@{\extracolsep{\fill}}lcccc@{}}
\toprule
Setting & $\Delta$PSNR & LPIPS W/L & DISTS W/L & Verdict \\
\midrule
Dense & -- & -- & -- & -- \\
Default & \pd{+7.23}{+6.60}{+7.85}{32/0} & 32/0 & 31/1 & win \\
Default, density cutoff & \pd{+5.98}{+5.46}{+6.50}{32/0} & 32/0 & 30/2 & win \\
HEART (EBC) & \pd{+6.96}{+6.35}{+7.56}{32/0} & 32/0 & 31/1 & win \\
HEART (EBC + TMR) & \pd{+6.59}{+5.99}{+7.19}{32/0} & 32/0 & 31/1 & win \\
Efficient-vDiT & \pd{+7.31}{+6.74}{+7.88}{32/0} & 32/0 & 32/0 & win \\
DiTFastAttn-style~\cite{ditfastattn} & \pd{+7.01}{+6.55}{+7.47}{32/0} & 32/0 & 32/0 & win \\
ScalingAttention (FAST) & \pd{+3.63}{+3.04}{+4.23}{31/1} & 31/1 & 30/2 & win \\
PASA & \pd{+4.47}{+4.14}{+4.81}{32/0} & 32/0 & 32/0 & win \\
CalibAtt-style & \pd{+5.69}{+5.17}{+6.21}{32/0} & 32/0 & 30/2 & win \\
DFSAttn-style~\cite{dfsattn} & \pd{+1.53}{+1.26}{+1.80}{31/1} & 30/2 & 28/4 & win \\
ALLOT & -- & -- & -- & -- \\
ALLOT + TMR & -- & -- & -- & -- \\
ALLOT (default time) & -- & -- & -- & -- \\
\bottomrule
\end{tabular*}
\end{table*}

\paragraph{VBench.}
The VBench column of Table~\ref{tab:compare} is the weighted composite of the 6 dimensions used by HEART, with VBench's official normalization and weights.
ALLOT is \pd{+0.30}{+0.004}{+0.59}{22/10} over the original FAST and \pd{+0.13}{-0.11}{+0.38}{20/12} over the original PASA, \pd{+0.18}{+0.09}{+0.27}{24/8} over the DFSAttn-style rule, and ALLOT + TMR is \dci{-0.28}{-1.19}{+0.62} relative to HEART (EBC + TMR); all remaining intervals contain zero except DiTFastAttn-style (\dci{-0.89}{-1.76}{-0.02}).

\paragraph{Held-out set.}
Rerunning these settings except TMR on 32 held-out prompts leaves every verdict unchanged. ALLOT's PSNR gains over EBC, Efficient-vDiT, the default and the CalibAtt-style rule range from 5.81 to 7.33\,dB (all 32/0). Its PSNR gains in dB are \pd{+3.76}{+2.96}{+4.55}{31/1} over FAST, \pd{+4.57}{+4.11}{+5.03}{32/0} over PASA, and \pd{+1.48}{+1.20}{+1.75}{31/1} over the DFSAttn-style rule. LPIPS and DISTS are better in every row (W/L 28/4--32/0).

\subsection{XAttention on HunyuanVideo}
\label{app:G4}

\paragraph{Setting.}
This block follows the HunyuanVideo setting of HEART's Table~2, also used by ScalingAttention (HunyuanVideo-T2V, 480p$\times$129 with frames of 720$\times$480, 50 steps, $n{=}32$).
Selection and kernel are those of \S\ref{app:G2}, applied to the concatenated [video $|$ text] sequence, and all settings keep the first 5 steps dense.
EBC is calibrated as in \S\ref{app:G2} on 10 HunyuanVideo prompts that do not overlap with the evaluation set.
FAST follows \S\ref{app:G2} on the same 10 calibration trajectories (selected $\gamma{=}4$, $\omega{=}2$) and, as on Wan, uses XAttention's masks rather than ScalingAttention's static topology and kernel.
ALLOT's table is solved from HunyuanVideo's mean-pooled probes, the same trajectory gain as \S\ref{app:J1} and the latency table of this XAttention path, targeting the default's wall-clock time.

\paragraph{Results (Table~\ref{tab:G4}).}
ALLOT runs at the default's wall-clock time (481.4 / 481.3\,s) and is 0.3\% and 2.6\% faster than EBC and FAST.
In PSNR it is \pd{+11.84}{+10.85}{+12.83}{32/0}, \pd{+10.88}{+9.87}{+11.88}{32/0} and \pd{+7.16}{+6.20}{+8.12}{32/0} above the default, EBC and FAST. SSIM, LPIPS and DISTS are better on all 32 prompts in all three comparisons.
On VBench (same accounting as \S\ref{app:G2}) it is \dci{+4.34}{+2.40}{+6.28}, \dci{+2.75}{+1.15}{+4.36} and \dci{+2.24}{+1.44}{+3.03} above the three.

\begin{table}[t]
\centering\ttiny
\caption{XAttention on HunyuanVideo 480p$\times$129 (720$\times$480; $n{=}32$; speedup = dense wall-clock / row wall-clock). Paired differences in \S\ref{app:G4}.}
\label{tab:G4}
\resizebox{\columnwidth}{!}{%
\begin{tabular}{lcccccc}
\toprule
Allocation & PSNR$\uparrow$ & SSIM$\uparrow$ & LPIPS$\downarrow$ & VBench$\uparrow$ & Wall-clock (s) & Speedup \\
\midrule
Dense & --- & --- & --- & 70.99 & 595.4 & 1.00$\times$ \\
Default (threshold 0.90) & 15.65 & 0.568 & 0.476 & 65.62 & 481.3 & 1.24$\times$ \\
HEART (EBC) & 16.61 & 0.608 & 0.428 & 67.20 & 483.0 & 1.23$\times$ \\
ScalingAttention (FAST) & 20.33 & 0.705 & 0.328 & 67.72 & 494.4 & 1.20$\times$ \\
ALLOT & \textbf{27.49} & \textbf{0.875} & \textbf{0.122} & 69.96 & 481.4 & 1.24$\times$ \\
\bottomrule
\end{tabular}}
\end{table}

\subsection{XAttention on 1.3B \texorpdfstring{720p$\times$81}{720px81}}
\label{app:G5}

\paragraph{Setting.}
This block follows the Wan2.1-1.3B 1280$\times$720$\times$81 setting of HEART's Table~2 (50 steps, CFG 5, UniPC, $n{=}32$). We use the model's default flow shift of 3.0, as in the Radial and SVOO 1.3B 720p rows of Table~\ref{tab:plugin}.
Selection, kernel, the 5 dense leading steps and TMR are those of \S\ref{app:G2}.
EBC's per-head thresholds and ALLOT's cell probes and trajectory gain all reuse the 480p$\times$81 calibration. The latency table is measured at 720p.
ALLOT's table is solved targeting the default's wall-clock time (421 dense cells), and ALLOT + TMR applies TMR to the same table.

\paragraph{Results (Table~\ref{tab:G5}; paired intervals in Table~\ref{tab:G1}).}
ALLOT is 1.2\% and 1.1\% faster than the default and EBC, and ALLOT + TMR runs at the wall-clock time of HEART (EBC + TMR) (\S\ref{app:A2}).
In PSNR it is 6.35, 5.16 and 5.21\,dB above the default, EBC and HEART (EBC + TMR). SSIM is better on all 32 prompts in all three comparisons and LPIPS wins 32/0, 31/1 and 31/1.
On VBench (same accounting as \S\ref{app:G2}) it is \dci{+0.99}{-0.20}{+2.18}, \dci{+1.02}{+0.17}{+1.86} and \dci{+1.30}{+0.21}{+2.38} above the three.

\begin{table}[t]
\centering\ttiny
\caption{XAttention on Wan2.1-1.3B 720p$\times$81 ($n{=}32$).}
\label{tab:G5}
\resizebox{\columnwidth}{!}{%
\begin{tabular}{lccccc}
\toprule
Allocation & PSNR$\uparrow$ & SSIM$\uparrow$ & LPIPS$\downarrow$ & VBench$\uparrow$ & Wall-clock (s) \\
\midrule
Dense & --- & --- & --- & 71.12 & 715.9 \\
Default (threshold 0.90) & 21.16 & 0.731 & 0.287 & 71.08 & 499.3 \\
HEART (EBC) & 22.36 & 0.762 & 0.241 & 71.05 & 498.5 \\
ALLOT & \textbf{27.51} & \textbf{0.841} & \textbf{0.146} & 72.07 & 493.2 \\
HEART (EBC + TMR) & 22.17 & 0.751 & 0.252 & 70.73 & 441.5 \\
ALLOT + TMR & \textbf{27.38} & \textbf{0.835} & \textbf{0.152} & 72.02 & 442.3 \\
\bottomrule
\end{tabular}}
\end{table}

\section{Published Pipelines and Fidelity--Speedup Curves}
\label{app:H}

Speedups in Table~\ref{tab:plugin} are relative to the dense wall-clock times in Table~\ref{tab:Hdense}: the dense path of the same kernel for SpargeAttn and Radial, and the dense runs of their own pipelines for SVOO and SVG2 (\S\ref{app:Hsvoo} and \S\ref{app:H3}).

\begin{table}[t]
\centering\ttiny
\caption{Dense wall-clock times (s) used for the speedups of Table~\ref{tab:plugin}. Wan models at 720p$\times$81 except SpargeAttn (480p$\times$81); HunyuanVideo at 720p$\times$129.}
\label{tab:Hdense}
\resizebox{\columnwidth}{!}{%
\begin{tabular}{@{}lcccc@{}}
\toprule
Pipeline & Wan2.1-1.3B & Wan2.1-14B & Wan2.2-A14B & HunyuanVideo \\
\midrule
SpargeAttn & 183.8 & -- & -- & -- \\
Radial & 728.1 & 3576.1 & -- & 3192.3 \\
SVOO & 700.1 & 3406.0 & 3355.9 & 3225.5 \\
SVG2 & 682.9 & -- & -- & -- \\
\bottomrule
\end{tabular}}
\end{table}

\subsection{SVOO}
\label{app:Hsvoo}

\paragraph{Protocol.}
SVOO's official repository~\cite{svoo} is run unchanged, with hyperparameters copied from its 720p scripts.

\paragraph{14B 720p row.}
The table is solved from probes under SVOO's selection strategy and the latency table of the SVOO path (\S\ref{app:F4}), reaching equal wall-clock time with the default (\S\ref{app:A2}; 2586.6 / 2603.5\,s): PSNR 31.70\,dB, \pd{+3.55}{+2.95}{+4.15}{32/0}, and SSIM, LPIPS and DISTS are all better as well (Table~\ref{tab:Hsvoo}).

\paragraph{1.3B 720p row.}
The table is solved with the latency table of the SVOO path (\S\ref{app:F4}), reaching equal wall-clock time with the default (447.7 / 451.6\,s): PSNR 30.73\,dB, \pd{+0.66}{+0.19}{+1.13}{21/11}, but LPIPS (9/23) and DISTS (4/28) favor the default (Table~\ref{tab:Hsvoo}).

\paragraph{HunyuanVideo 720p$\times$129 row.}
The setting follows the main table of the SVOO paper (HunyuanVideo-T2V, 129 frames at a resolution of 720$\times$1280, 50 steps), and the default copies SVOO's official HunyuanVideo configuration.
The table reuses the cell probes measured under SVOO's selection strategy on HunyuanVideo 480p and the trajectory gain of \S\ref{app:J1}. Only the latency table changes; as for 14B, it is fitted to the end-to-end wall-clock time of four uniform densities on the SVOO path. It targets the default's wall-clock time.
ALLOT is 7.9\% faster than the default (1697.0 / 1842.7\,s). The dense wall-clock time is the median over prompts 0 and 1 (3225.5\,s).
PSNR 30.93\,dB, \pd{+3.38}{+2.51}{+4.25}{31/1}, and SSIM, LPIPS and DISTS are all better as well (Table~\ref{tab:Hsvoo}).

\paragraph{VBench.}
With the same accounting as \S\ref{app:G2}, the per-prompt paired differences ALLOT $-$ default are: 14B 720p \dci{-0.08}{-0.29}{+0.12}, Wan2.2 720p \dci{-0.32}{-0.94}{+0.29}, 1.3B 720p \dci{-0.66}{-1.55}{+0.22}, HunyuanVideo 720p$\times$129 \dci{-0.25}{-1.04}{+0.55}; none of the four is significant.

\begin{table*}[t]
\centering\readablepairedtable
\caption{Paired differences (ALLOT $-$ default) for the SVOO rows of Table~\ref{tab:plugin}; verdicts follow \S\ref{app:A1}.}
\label{tab:Hsvoo}
\begin{tabular*}{\textwidth}{@{\extracolsep{\fill}}P{2.8cm}P{2.2cm}cccccc@{}}
\toprule
Setting & Baseline & $n$ & $\Delta$PSNR & $\Delta$SSIM & $\Delta$LPIPS & $\Delta$DISTS & Verdict \\
\midrule
SVOO 14B 720p & default & 32 & \pd{+3.55}{+2.95}{+4.15}{32/0} & \pd{+0.027}{+0.019}{+0.035}{31/1} & \pd{-0.024}{-0.031}{-0.017}{31/1} & \pd{-0.011}{-0.013}{-0.008}{30/2} & win \\
SVOO Wan2.2 720p & default & 32 & \pd{+4.87}{+4.37}{+5.38}{32/0} & \pd{+0.057}{+0.042}{+0.072}{32/0} & \pd{-0.045}{-0.055}{-0.035}{32/0} & \pd{-0.014}{-0.016}{-0.012}{32/0} & win \\
SVOO 1.3B 720p & default & 32 & \pd{+0.66}{+0.19}{+1.13}{21/11} & \pd{-0.003}{-0.009}{+0.004}{14/18} & \pd{+0.012}{+0.004}{+0.021}{9/23} & \pd{+0.017}{+0.011}{+0.023}{4/28} & tie \\
SVOO HunyuanVideo 720p$\times$129 & default & 32 & \pd{+3.38}{+2.51}{+4.25}{31/1} & \pd{+0.039}{+0.029}{+0.050}{30/2} & \pd{-0.047}{-0.060}{-0.035}{29/3} & \pd{-0.019}{-0.023}{-0.014}{29/3} & win \\
\bottomrule
\end{tabular*}
\end{table*}

\subsection{SpargeAttn}
\label{app:H1}

SpargeAttn's released tuned hyperparameters~\cite{spargeattn} are for Wan2.1-1.3B at 480p$\times$81, the setting of Table~\ref{tab:plugin}.
At 720p$\times$81 we run both the official checkpoint and one retuned on our grid with its built-in tuner: PSNR 28.64 $\to$ 30.88 and 29.30 $\to$ 31.29\,dB (520.6 $\to$ 514.2 and 574.4 $\to$ 527.5\,s; speedup 1.40$\times$ $\to$ 1.42$\times$ and 1.27$\times$ $\to$ 1.38$\times$).
All its default settings add a 30\% warm-up, and warm-up steps use native SDPA, the same as the dense cells of ALLOT's tables.
In ALLOT's SpargeAttn tables 40--55\% of the cells are dense and use SDPA.

\begin{table*}[t]
\centering\readablepairedtable
\caption{Paired differences (ALLOT $-$ default) for the SpargeAttn rows; verdicts follow \S\ref{app:A1}. The first row is from Table~\ref{tab:plugin}, the other two are the 720p settings above.}
\label{tab:H1}
\begin{tabular*}{\textwidth}{@{\extracolsep{\fill}}P{2.8cm}P{1.2cm}cccccc@{}}
\toprule
Setting & Baseline & $n$ & $\Delta$PSNR & $\Delta$SSIM & $\Delta$LPIPS & $\Delta$DISTS & Verdict \\
\midrule
SpargeAttn 1.3B 480p$\times$81, official configuration & default & 32 & \pd{+5.03}{+4.49}{+5.57}{32/0} & \pd{+0.108}{+0.083}{+0.133}{32/0} & \pd{-0.080}{-0.100}{-0.061}{32/0} & \pd{-0.041}{-0.051}{-0.030}{32/0} & win \\
SpargeAttn 1.3B 720p, official configuration & default & 32 & \pd{+2.24}{+1.82}{+2.65}{31/1} & \pd{+0.035}{+0.022}{+0.047}{29/3} & \pd{-0.032}{-0.044}{-0.019}{28/4} & \pd{-0.015}{-0.025}{-0.005}{23/9} & win \\
SpargeAttn 1.3B 720p, retuned & default & 32 & \pd{+1.99}{+1.63}{+2.36}{31/1} & \pd{+0.028}{+0.016}{+0.041}{30/2} & \pd{-0.029}{-0.041}{-0.018}{28/4} & \pd{-0.014}{-0.022}{-0.006}{27/5} & win \\
\bottomrule
\end{tabular*}
\end{table*}

\subsection{Radial}
\label{app:H2}

Radial's mask code and kernel (FlashInfer fixed 128-block BSR)~\cite{radial} are used as is.
Radial's official Wan setting uses decay 0.2 (Table~\ref{tab:plugin}). A second operating point with decay 1.0 gives 28.45 $\to$ 30.45\,dB at equal wall-clock time (500.9 / 499.8\,s; Table~\ref{tab:H2}).

At 480p$\times$33 the minimum density of the mask is 0.525, so on Wan, Radial is run only at 720p$\times$81, and the two groups differ only in each cell's \texttt{decay\_factor}.
The table of the 14B row is solved from mean-pooled probes.

\paragraph{HunyuanVideo 720p$\times$129 row.}
The default is Radial's official HunyuanVideo configuration (first 12 steps dense, no dense layer, decay 0.95). We generate 129 frames at a resolution of $720{\times}1280$.
ALLOT's table is solved from probes measured under Radial (\S\ref{app:F2}), with \pd{-0.55}{-1.17}{+0.07}{16/16} at equal wall-clock time (1805.2 / 1814.2\,s). The SSIM, LPIPS and DISTS intervals all contain 0.

\begin{table*}[t]
\centering\readablepairedtable
\caption{Paired differences (ALLOT $-$ default) for the Radial rows of Table~\ref{tab:plugin} and the decay 1.0 setting; verdicts follow \S\ref{app:A1}.}
\label{tab:H2}
\begin{tabular*}{\textwidth}{@{\extracolsep{\fill}}P{2.8cm}P{1.2cm}cccccc@{}}
\toprule
Setting & Baseline & $n$ & $\Delta$PSNR & $\Delta$SSIM & $\Delta$LPIPS & $\Delta$DISTS & Verdict \\
\midrule
Radial 1.3B 720p, decay 0.2 & default & 32 & \pd{+0.40}{+0.11}{+0.69}{26/6} & \pd{+0.005}{+0.000}{+0.010}{24/8} & \pd{-0.005}{-0.013}{+0.002}{23/9} & \pd{-0.004}{-0.010}{+0.003}{21/11} & win \\
Radial 1.3B 720p, decay 1.0 & default & 32 & \pd{+2.00}{+1.46}{+2.54}{30/2} & \pd{+0.016}{+0.009}{+0.024}{26/6} & \pd{-0.012}{-0.021}{-0.004}{25/7} & \pd{-0.007}{-0.013}{+0.000}{23/9} & win \\
Radial 14B 720p & default & 32 & \pd{+0.53}{+0.37}{+0.70}{28/4} & \pd{+0.011}{+0.006}{+0.015}{28/4} & \pd{-0.013}{-0.017}{-0.009}{28/4} & \pd{-0.007}{-0.011}{-0.003}{29/3} & win \\
Radial HunyuanVideo 720p$\times$129 & default & 32 & \pd{-0.55}{-1.17}{+0.07}{16/16} & \pd{-0.000}{-0.008}{+0.007}{18/14} & \pd{+0.010}{-0.003}{+0.023}{12/20} & \pd{+0.003}{-0.004}{+0.010}{16/16} & tie \\
\bottomrule
\end{tabular*}
\end{table*}

\paragraph{VBench of the two pipelines.}
For VBench on these settings (same accounting as \S\ref{app:G2}), ALLOT is significantly higher on the two SpargeAttn official-configuration rows (\pd{+1.55}{+0.78}{+2.31}{27/5} and \pd{+1.19}{+0.14}{+2.24}{26/6}); the intervals of the other five rows all contain 0.

\subsection{SVG2}
\label{app:H3}

\paragraph{Protocol.}
SVG2's official Wan pipeline~\cite{svg2} is run (Wan2.1-1.3B, 1280$\times$720$\times$81, 50 steps, CFG 5, UniPC). SVOO, which builds on SVG2's code, covers the other three models (\S\ref{app:Hsvoo}).
Hyperparameters are copied from its 720p script (top-$p$ 0.90, minimum key-cluster ratio 0.10). Following HEART's protocol, the warm-up keeps the first 5 steps dense.
The reference is the same pipeline with full attention at every step.

\paragraph{Table.}
Cell probes are re-measured under SVG2's selection strategy, and the latency table is measured on the SVG2 path at this shape. The table is solved at the default's wall-clock time (410.5 / 410.9\,s).
The dense wall-clock time, 682.9\,s, is obtained from the dense-to-default time ratio, timed alternately on 6 prompts.

\paragraph{Results.}
PSNR 25.60\,dB, \pd{+1.34}{+0.64}{+2.05}{26/6}; the mean SSIM is slightly better (\pd{+0.012}{-0.001}{+0.024}{17/15}), while the mean LPIPS is slightly worse (\pd{+0.005}{-0.015}{+0.024}{13/19}) and DISTS favors the default (\pd{+0.021}{+0.004}{+0.037}{6/26}).

\paragraph{VBench.}
The per-prompt paired difference ALLOT $-$ default is \dci{-0.34}{-1.48}{+0.80}, not significant.

\subsection{Fidelity--speedup curves at \texorpdfstring{720p$\times$81}{720px81}}
\label{app:H4}

The budget is swept, one table is solved per budget and paired with a uniform baseline at matched wall-clock time. The wall-clock time of the two groups differs by $\le$2\%.

\begin{figure*}[t]
\centering
\includegraphics[width=\textwidth]{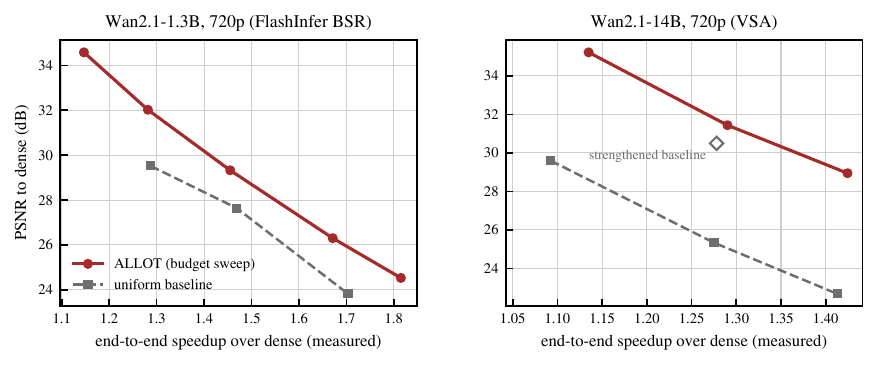}
\caption{Fidelity--speedup curves at 720p$\times$81 (left Wan2.1-1.3B, right Wan2.1-14B). Red: ALLOT; gray: uniform baseline at equal wall-clock time; diamonds: strengthened baseline.}
\label{fig:H1}
\end{figure*}

\section{Ablations}
\label{app:I}

\subsection{Ablation on XAttention}
\label{app:I1}

On XAttention, the parametric shape rule of Table~\ref{tab:ablation} takes its shape from the table solved on XAttention: that table's first 10 steps are fully dense, layer~0 is dense in 74\% of the steps, and every other layer in less than 50\%. The rule therefore keeps the first 10 steps and layer~0 dense over all steps, and lowers every other layer linearly with the step from 0.59 at step 10 to 0.05 at step 49, with the coefficient set by bisection on the latency table to match ALLOT's cost.

\begin{table*}[t]
\centering\readablepairedtable
\caption{Paired differences (row $-$ ALLOT) for the rows of Table~\ref{tab:ablation} (ablation, no verdict).}
\label{tab:I1}
\begin{tabular*}{\textwidth}{@{\extracolsep{\fill}}P{2.5cm}P{2.8cm}ccccc@{}}
\toprule
Setting & Baseline & $n$ & $\Delta$PSNR & $\Delta$SSIM & $\Delta$LPIPS & $\Delta$DISTS \\
\midrule
XAttention 1.3B 480p$\times$81 & same density for all layers (trajectory gain only) & 32 & \pd{-1.07}{-1.34}{-0.80}{2/30} & \pd{-0.023}{-0.029}{-0.018}{1/31} & \pd{+0.022}{+0.017}{+0.027}{2/30} & \pd{+0.021}{+0.016}{+0.026}{3/29} \\
XAttention 1.3B 480p$\times$81 & trajectory gain all 1 & 32 & \pd{-7.05}{-7.52}{-6.58}{0/32} & \pd{-0.148}{-0.173}{-0.123}{0/32} & \pd{+0.131}{+0.110}{+0.152}{0/32} & \pd{+0.042}{+0.033}{+0.051}{1/31} \\
XAttention 1.3B 480p$\times$81 & parametric shape rule & 32 & \pd{-1.76}{-2.02}{-1.50}{1/31} & \pd{-0.028}{-0.034}{-0.023}{1/31} & \pd{+0.023}{+0.018}{+0.028}{1/31} & \pd{+0.014}{+0.010}{+0.018}{3/29} \\
XAttention 1.3B 480p$\times$81 & density-proportional cost & 32 & \pd{-0.12}{-0.26}{+0.03}{14/18} & \pd{+0.002}{-0.000}{+0.005}{20/12} & \pd{-0.001}{-0.003}{+0.001}{16/16} & \pd{-0.002}{-0.004}{-0.000}{20/12} \\
\bottomrule
\end{tabular*}
\end{table*}

\subsection{Density-proportional cost}
\label{app:I2}

On the setting of Table~\ref{tab:ablation} (XAttention, 1.3B 480p$\times$81, $n{=}32$), we keep everything else in ALLOT fixed, only change the cost in the solver to $\lut(\dens)=\dens$, and then align the budget to ALLOT's cost using the latency table.
The resulting table has mean density 0.48, yet its latency-table cost is 0.73 of dense. It has 446 dense cells (ALLOT 587), and its PSNR, SSIM and LPIPS do not differ significantly from ALLOT's (Table~\ref{tab:I1}).

\subsection{Calibration metric of the trajectory gain}
\label{app:I3}

\paragraph{Setting.}
1.3B 720p$\times$81, SVOO pipeline, $n{=}32$.
The pixel-MSE calibration uses the table at a budget of 0.62 of dense: probes under SVOO's selection strategy and the latency table of the SVOO path (\S\ref{app:F4}). The LPIPS and DISTS calibrations differ only in the metric used to aggregate $M_k$.

\paragraph{Results.}
The pixel-MSE table gives 31.12\,dB. Calibrating with LPIPS instead gives 30.28\,dB (\pd{-0.84}{-1.20}{-0.49}{4/28}); calibrating with DISTS gives 29.06\,dB (\pd{-2.06}{-2.71}{-1.41}{1/31}). Both alternatives also worsen LPIPS itself.
Perceptual metrics saturate at large errors (LPIPS already reaches 0.52 in the first window, while MSE differs by about 460$\times$), so they underweight early steps: the ratio of $w$ over the first and last 10 steps is only 1/9 (LPIPS) and 1/29 (DISTS) of that under MSE calibration, and the densities of the early steps drop accordingly (dense cells 617 $\to$ 590 / 532).

\subsection{Ablation on mean-pooled top-$k$}
\label{app:I4}

Every row is on 1.3B 480p$\times$33 with the same selection strategy (mean-pooled top-$k$), kernel (VSA) and budget, $n{=}32$, replacing one module at a time with the rest unchanged.

\begin{table*}[t]
\centering\ttiny
\caption{Ablation on mean-pooled top-$k$ ($\Delta$ = row $-$ ALLOT).}
\label{tab:I2}
\suppfit{\begin{tabular}{@{}lcccc@{}}
\toprule
Setting & PSNR / LPIPS / DISTS & $\Delta$PSNR & LPIPS W/L & DISTS W/L \\
\midrule
ALLOT & 25.99 / 0.098 / 0.107 & -- & -- & -- \\
Same density for all layers (trajectory gain only) & 22.77 / 0.189 / 0.177 & \pd{-3.22}{-3.78}{-2.67}{0/32} & 0/32 & 0/32 \\
Trajectory gain all 1 & 14.33 / 0.389 / 0.187 & \pd{-11.66}{-13.17}{-10.16}{0/32} & 0/32 & 2/30 \\
Parametric shape rule & 24.86 / 0.121 / 0.129 & \pd{-1.13}{-1.27}{-0.99}{0/32} & 0/32 & 0/32 \\
\bottomrule
\end{tabular}}
\end{table*}

As on XAttention (Table~\ref{tab:I1}), the largest contributor is trajectory calibration.

\section{Cross-Model Comparison}
\label{app:J}

All Wan settings of this section use mean-pooled top-$k$ selection and the VSA kernel (the 1.3B curve of \S\ref{app:H4} uses FlashInfer BSR), and settings differ only in the density table.
\begin{itemize}[nosep,leftmargin=1.5em]
\item \textbf{Uniform baseline:} first 15 steps dense, all other cells at one density.
\item \textbf{Strengthened baseline:} layer~0 dense over all steps + first 20 steps dense + uniform elsewhere (20 steps is the best of $\{0,5,10,15,20,25\}$ on 1.3B; 14B uses 20).
\item \textbf{Parametric shape rule} (Table~\ref{tab:I2}): first 20 steps and layers 0/1 dense over all steps, other layers decreasing linearly with the step, shape taken from the solved table.
\end{itemize}

\subsection{Summary of four models}
\label{app:J1}

HunyuanVideo uses FlexAttention~\cite{flexattention} and is compared at equal density ($\lut(\dens)=\dens$): at this shape the FlexAttention path offers little or no speedup over dense attention, so an equal-latency budget would leave almost nothing to allocate.
On 32 held-out prompts, 1.3B is \pd{+2.18}{+1.84}{+2.51}{32/0} over the strengthened baseline and \pd{+6.49}{+5.62}{+7.35}{32/0} over the uniform baseline; 14B 480p is \pd{+0.44}{+0.22}{+0.66}{26/6} over the strengthened baseline (DISTS 12/20, not significant) and \pd{+8.82}{+8.07}{+9.57}{32/0} over the uniform baseline.

\begin{table*}[t]
\centering\ttiny
\caption{PSNR (dB) of four models under the same budget. $^*$:~the uniform baseline with layer~0 additionally dense over all steps (the number of warm-up steps was not searched).}
\label{tab:J1}
\suppfit{\begin{tabular}{@{}lccccccc@{}}
\toprule
Model and setting & Speedup & $n$ & Uniform baseline & Strengthened baseline & ALLOT & ALLOT $-$ strengthened baseline & ALLOT $-$ uniform baseline \\
\midrule
Wan2.1-14B, 720p$\times$81 & 1.29$\times$ & 16 & 24.87 & 30.49 & 31.25 & \pd{+0.76}{+0.46}{+1.06}{15/1} & \pd{+6.38}{+5.55}{+7.21}{16/0} \\
Wan2.1-1.3B, 480p$\times$33 & 1.14$\times$ & 32 & 19.59 & 23.44 & 25.99 & \pd{+2.55}{+2.16}{+2.94}{32/0} & \pd{+6.40}{+5.63}{+7.17}{32/0} \\
Wan2.1-14B, 480p$\times$33 & 1.09--1.11$\times$ & 32 & 19.20 & 27.26 & 27.85 & \pd{+0.60}{+0.41}{+0.79}{28/4} & \pd{+8.65}{+7.84}{+9.46}{32/0} \\
Wan2.2-A14B, 480p$\times$33 & 1.09$\times$ & 32 & 19.36 & 24.20$^*$ & 25.86 & \pd{+1.67}{+1.37}{+1.96}{32/0}$^*$ & \pd{+6.50}{+5.92}{+7.07}{32/0} \\
HunyuanVideo, 480p$\times$33 & equal density & 32 & 27.82 & 27.46 & 28.42 & \pd{+0.96}{+0.02}{+1.90}{24/8} & \pd{+0.60}{-0.34}{+1.53}{22/10} \\
\bottomrule
\end{tabular}}
\end{table*}

\begin{table*}[t]
\centering\ttiny
\caption{PSNR (dB, $n{=}32$) of the strengthened baseline as a function of $k$ on Wan2.1-1.3B 480p$\times$33, and ALLOT.}
\label{tab:J2}
\suppfit{\begin{tabular}{@{}lcccccccccc@{}}
\toprule
Budget & $k{=}5$ & 10 & 15 & 20 & 25 & 30 & 35 & ALLOT & ALLOT $-$ best $k$ & ALLOT $-$ $k{=}20$ \\
\midrule
0.20 & 16.92 & 19.73 & \textbf{20.23} & 19.13 & 15.84 & -- & -- & 22.79 & \pd{+2.56}{+2.18}{+2.94}{32/0} & \pd{+3.66}{+3.26}{+4.06}{32/0} \\
0.30 & 17.85 & 21.00 & 22.71 & \textbf{23.44} & 23.14 & -- & -- & 25.99 & \pd{+2.55}{+2.16}{+2.94}{32/0} & \pd{+2.55}{+2.16}{+2.94}{32/0} \\
0.45 & -- & -- & 25.15 & 26.74 & 27.77 & 28.60 & \textbf{29.05} & 31.15 & \pd{+2.09}{+1.73}{+2.45}{32/0} & \pd{+4.41}{+3.94}{+4.88}{32/0} \\
\bottomrule
\end{tabular}}
\end{table*}

\begin{table*}[t]
\setcounter{table}{0}\renewcommand{\thetable}{K.\arabic{table}}\renewcommand{\theHtable}{K.\arabic{table}}
\centering\ttiny
\caption{Robustness of the measurement setup. Agreement: share of cells whose dense/sparse status matches the original table (\mbox{Wan2.1-1.3B} unless noted).}
\label{tab:K1}
\begin{tabular}{@{}llP{8.4cm}l@{}}
\toprule
Variable & Change & Agreement (\%) & Correlation of $w$'s shape \\
\midrule
Probing videos & 4 $\to$ 1 & XAttention 96.1--98.5; mean-pooled: 1.3B 95.2--98.0, 14B 97.9--98.4, HunyuanVideo 93.4--96.3 & -- \\
Probing videos & 4 $\to$ 2 & XAttention 95.7--99.3; mean-pooled: 1.3B 96.7--97.5, 14B 98.5--99.4, HunyuanVideo 96.7--97.7 & -- \\
Calibration prompts & 8 $\to$ 1 & XAttention 92.3--98.5; mean-pooled: 1.3B 89.1--98.5, HunyuanVideo 86.1--97.9 & -- \\
Calibration prompts & 8 $\to$ 4 & 96.7--99.4 & -- \\
Window density & 0.30 $\to$ 0.15 / 0.50 & XAttention 98.9 / 98.5; mean-pooled 99.6 / 97.1 & 0.995 / 0.995 \\
Window selection strategy & mean-pooled $\to$ SpargeAttn & 96.5 / 96.0 (two budgets) & 0.993 \\
\bottomrule
\end{tabular}
\end{table*}

\subsection{Wan2.2-A14B}
\label{app:J2}

The two 14B experts split the work at $t{=}0.875$ and share one table, and the probe noise levels $t$ are placed on each expert's own schedule steps. The table is solved with 14B's latency table and Wan2.2's own measured trajectory gain, and the three groups use the same layer-step budget.

\subsection{Solved tables}
\label{app:J3}

The solved warm-up (the number of consecutive fully dense steps from step~0) grows as the budget increases: 10 / 15 / 20 steps at the three 14B 720p budgets (15 dense steps + uniform 0.20 / 0.30 / 0.45), and 5--20 steps at the five 1.3B 720p budgets.
Under the same budget, the solved warm-up is 15 steps on the Wan models, while on HunyuanVideo no step is fully dense: one of the 60 layers is at 0.30 from step~0 and four are sparse in steps 5--9. Layers 0 and 1 are dense in all 50 steps on HunyuanVideo and in 47 steps or more on Wan.

\subsection{Strengthened baseline at other budgets}
\label{app:J4}

On 1.3B 480p$\times$33, we apply the form of the strengthened baseline (layer~0 dense over all steps + the first $k$ steps dense + one density for the other layers) at two other budgets and re-search $k$.
Each budget is the cost under the same latency table of ``first 15 steps dense + uniform $\dens$'', with $\dens = 0.20$ / $0.45$ (the same as \S\ref{app:C4}). For each $k$ the density of the other layers is set via the latency table to reach the same cost.
At each budget ALLOT is re-solved with the same cell probes and trajectory gain.

The best $k$ changes with the budget: the $k{=}20$ selected at 0.30, moved to 0.20, is 1.10\,dB lower than the best $k$ of that budget ([0.61, 1.59], 29/3), and moved to 0.45 it is 2.32\,dB lower ([1.79, 2.85], 28/4). At 0.45 PSNR rises monotonically with $k$; at $k{=}40$ the density of the other layers would be only 0.01, so the search stops at $k{=}35$. After re-searching $k$, ALLOT is still a win at both new budgets (LPIPS 31/1 and 32/0, DISTS 21/11 and 32/0). All $k$ are selected on the evaluation set. The warm-up solved by ALLOT is 10 / 15 / 20 steps at the three budgets.

\section{Robustness of the Measurement Setup}
\label{app:K}

\subsection{Number of calibration inputs}
\label{app:K1}

Each variant re-solves the table and compares it with the 4-video, 8-prompt baseline (Table~\ref{tab:K1}).
With 2 probing videos, 96--99\% of cells keep the same dense/sparse status across models. On XAttention (1.3B), 1 or 2 videos lose less than 0.6\,dB, about 8\% of ALLOT's gain; 8 videos bring no significant improvement.
Step-wise rankings are more sensitive to calibration inputs than layer-wise rankings: the single-prompt ratio of $w$ over the first and last 10 steps ranges from 162 to 1507 on 1.3B (mean of 8: 333).

\subsection{Robustness of the trajectory gain}
\label{app:K2}

All selection strategies reuse the trajectory gain measured with mean-pooled top-$k$ (Table~\ref{tab:K1}).
Window density changes the magnitude of $w$ but barely its shape; calibrating separately at three densities changes 0.6\% of the cells and PSNR by \pd{-0.02}{-0.10}{+0.06}{15/17}.
Re-calibrating $w$ on 8 held-out prompts gives a log correlation of 0.999 with the original. When the 12 tables are re-solved with this $w$, 98.4--99.3\% of the cells retain their dense/sparse status.

\subsection{Probe resolution}
\label{app:K3}

All 81-frame tables (720p and 480p$\times$81) are solved from 480p$\times$33 probes, except Radial 1.3B which uses 720p latent probes (\S\ref{app:H2}).
SVOO's co-clustering changes with resolution, so on the SVOO pipeline (1.3B 720p$\times$81) the probes were re-measured from the dense 720p generations of 4 evaluation prompts and the table re-solved with the recipe of that row of Table~\ref{tab:plugin}. The mean final-video PSNR ($n{=}16$) increases from 30.43 to 30.48\,dB, \pd{+0.05}{-0.14}{+0.25}{9/7}, and the LPIPS and DISTS intervals both contain 0.

\subsection{Inference settings}
\label{app:K4}

The probes, trajectory gain and latency table are measured with 50 UniPC steps and CFG 5.0. On the first group of Table~\ref{tab:compare}, we change one inference setting at a time without re-measuring, and compare ALLOT with the default against a dense reference generated under the same setting (Table~\ref{tab:K2}).
With CFG 7.5 and the Euler sampler, ALLOT uses the table of Table~\ref{tab:compare} unchanged. With 30 steps, the table is re-solved on the 30-step schedule, taking each step's trajectory gain from the 50-step schedule at the nearest noise level.
ALLOT wins in all three settings at equal wall-clock time: \pd{+5.93}{+5.26}{+6.61}{32/0} with 30 steps, \pd{+6.80}{+6.26}{+7.34}{32/0} with CFG 7.5 and \pd{+7.18}{+6.41}{+7.95}{32/0} with the Euler sampler, compared with 7.23\,dB in the original setting.

\begin{table}[t]
\setcounter{table}{1}
\centering\footnotesize
\caption{Changing one inference setting after calibration (XAttention, Wan2.1-1.3B 480p$\times$81, $n{=}32$; other settings as in Table~\ref{tab:compare}). Fidelity is measured against the dense output under the same setting; wall-clock time follows Appendix~\ref{app:A}.}
\label{tab:K2}
\begin{tabular}{@{}llcccr@{}}
\toprule
Setting & Allocation & PSNR$\uparrow$ & SSIM$\uparrow$ & LPIPS$\downarrow$ & Wall-clock (s) \\
\midrule
30 steps & Default & 19.87 & 0.696 & 0.232 & 92.5 \\
 & ALLOT & \textbf{25.80} & \textbf{0.830} & \textbf{0.110} & 91.9 \\
\midrule
CFG 7.5 & Default & 17.68 & 0.668 & 0.255 & 150.1 \\
 & ALLOT & \textbf{24.48} & \textbf{0.827} & \textbf{0.101} & 147.7 \\
\midrule
Euler sampler & Default & 18.79 & 0.664 & 0.278 & 147.8 \\
 & ALLOT & \textbf{25.97} & \textbf{0.832} & \textbf{0.113} & 147.1 \\
\bottomrule
\end{tabular}
\end{table}

\end{document}